\documentclass[12pt]{article}
\pdfoutput=1
\usepackage{graphicx}
\usepackage{bm}
\usepackage{fancyhdr,graphicx,amsmath,amssymb}
\usepackage{subfigure}
\usepackage{xcolor}
\usepackage{diagbox}
\usepackage{multirow}
\usepackage[margin=0.5cm]{caption}
\usepackage{float}

\usepackage[ruled,vlined,linesnumbered]{algorithm2e}
\include{pythonlisting}
\usepackage{natbib} 
\usepackage{url} 

\newcommand{\blind}{0}

\newtheorem{prop}{Proposition}
\newtheorem{lemma}{Lemma}

\begin{document}

\def\spacingset#1{\renewcommand{\baselinestretch}%
{#1}\small\normalsize} \spacingset{1}


\if0\blind
{
  \title{\bf A Supervised Tensor Dimension Reduction-Based Prognostic Model for Applications with Incomplete Imaging Data}
  \author{Chengyu Zhou\\
    Edward P. Fitts Department of Industrial and Systems Engineering,\\ North Carolina State University
\\
    and \\
    Xiaolei Fang \\
    Edward P. Fitts Department of Industrial and Systems Engineering,\\ North Carolina State University
\\}
\date{}
  \maketitle
} \fi

\if1\blind
{
  \bigskip
  \bigskip
  \bigskip
  \begin{center}
    {\LARGE\bf Title}
\end{center}
  \medskip
} \fi

\bigskip
\begin{abstract}
Imaging data-based prognostic models focus on using an asset's degradation images to predict its time-to-failure (TTF). Most image-based prognostic models have two common limitations. First, they require degradation images to be complete (i.e., images are observed continuously and regularly over time). Second, they usually employ an unsupervised dimension reduction method to extract low-dimensional features, and then use the features for TTF prediction. Since unsupervised dimension reduction is conducted on the degradation images without the involvement of TTFs, there is no guarantee that the extracted features are effective for failure time prediction. To address these challenges, this article develops a supervised tensor dimension reduction-based prognostic model. The model first proposes a supervised dimension reduction method for tensor data. It uses historical TTFs to guide the detection of a tensor subspace to extract low-dimensional features from high-dimensional incomplete degradation imaging data. Next, the extracted features are used to construct a prognostic model based on (log)-location-scale regression. An optimization algorithm for parameter estimation is proposed and analytical solutions are discussed. Simulated data and a real-world data set are used to validate the performance of the proposed model.

\end{abstract}

\noindent%
\vfill

\newpage
\spacingset{1.5} 
\section{Introduction}
\label{sec:intro}

Degradation is an irreversible process of damage accumulation that results in the failure of engineering systems/assets/components \citep{bogdanoff1985probabilistic}. Although it is usually challenging to observe a physical degradation process, there often are some manifestations associated with degradation processes that can be monitored by sensing technology, which yields data known as degradation data/signals. Degradation data contains the health condition of engineering assets; thus, if modeled properly, they can be used to predict the assets' time-to-failure (TTF) via a process known as prognostic. Many prognostic models have been developed in the literature, most of which focus on using time series-based degradation data \citep{ye2014semiparametric,ye2014inverse,hong2010field,hong2013field,shu2015life,liu2013data,gebraeel2005residual,wang2022sgl}. Recently, prognostic models with imaging-based degradation data have been investigated and attracted more and more attention. This is because comparing with time-series data, imaging data usually contains much richer information of the object being monitored, and imaging sensing technologies are noncontact and thus they can usually be easily deployed. One example of imaging-based degradation data is the infrared image stream that measures the change of temperature distribution of a thrust bearing during its degradation process over time \citep{fang2019image,aydemir2019image,dong2021infrared,jiang2022adversarial,wang2021augmented}. Another example is the images used to measure the performance degradation  of infrared systems such as rotary-wing drones \citep{dong2021infrared}.

The existing imaging-based prognostic methods include deep learning-based models and statistical learning methods. Examples of the deep learning-based models designed for TTF prediction using imaging data include the ones developed by \cite{aydemir2019image,yang2021multi,dong2021infrared}, \cite{jiang2022adversarial}, \cite{jiang2022spatiotemporal}, and \cite{jiang2023sparse}. Although these models have worked relatively well, they usually provide point estimations of failure times, and it is challenging for them to quantify the uncertainty of predicted TTFs (e.g., providing a failure time distribution). This limits their applicability since the subsequent decision-making analysis such as maintenance/inventory/logistic optimization requires prognostic models to provide a distribution of the predicted TTF. Also, deep learning-based prognostic models often require a relatively large number of historical samples for model training, which cannot be satisfied by many real-world applications with limited historical data.  One example of statistical learning methods for image-based prognostic is the penalized (log)-location-scale (LLS) tensor regression proposed by \cite{fang2019image}. The model first employs multilinear principal component analysis (MPCA) \citep{lu2008mpca} to reduce the dimension of high-dimensional imaging-based degradation data, which yields a low-dimensional feature tensor for each asset. Next, it constructs a prognostic model by regressing an asset's TTF against its low-dimensional feature tensor using LLS regression. In the same article, \cite{fang2019image} have also proposed several benchmarking prognostic models that use imaging-based degradation data for TTF prediction. These models also first employ a dimension reduction method such as functional principal component analysis (FPCA) \citep{ramsay2005principal}, principal component analysis (PCA) \citep{abdi2010principal}, or B-Spline \citep{prautzsch2002bezier}, to reduce the dimension of degradation data and then use low-dimensional features to build an LLS regression model for prognostic. Although the aforementioned statistical learning-based prognostic models can provide a distribution for the predicted TTF, and their effectiveness have been well investigated, they share two common limitations.

\begin{figure}[htp!]
	\centering
	\subfigure[Complete Image Streams]{\raisebox{+0cm}{\includegraphics[width=6.5cm]{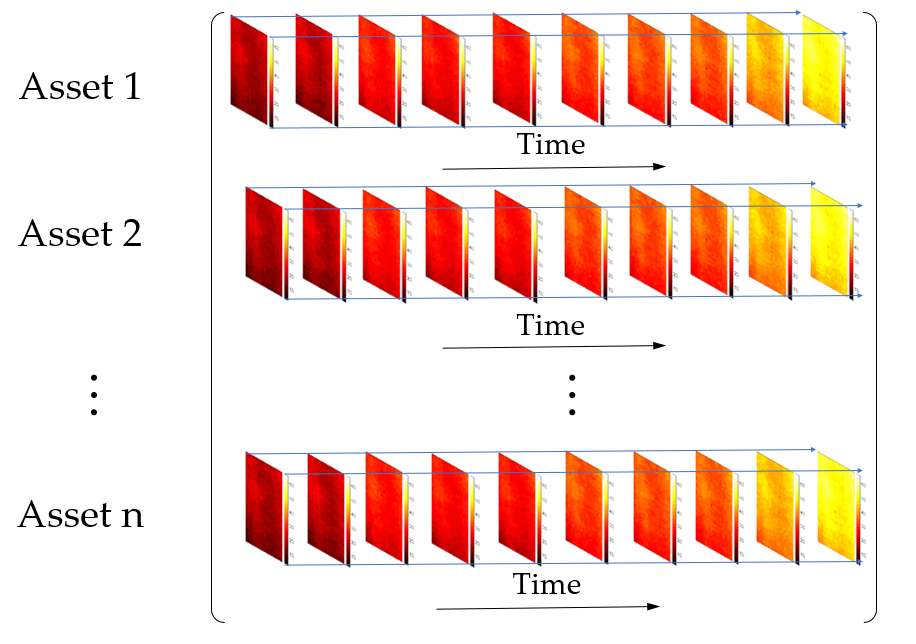} }}%
	\qquad \quad
	\subfigure[Incomplete Image Streams]{{\includegraphics[width=6.5cm]{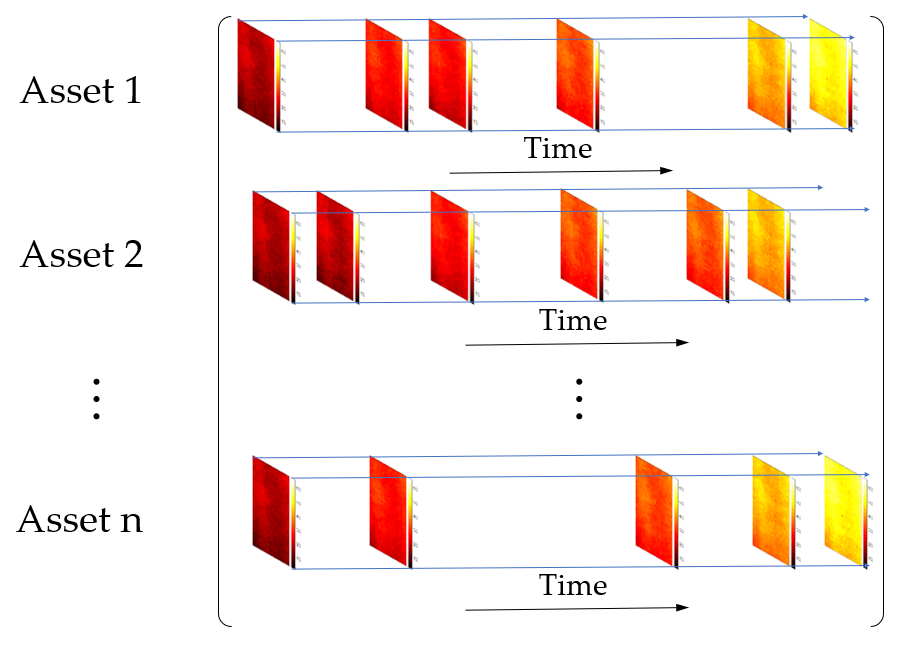} }}%
	\caption{ Degradation stream images with and without missing data.} %
	\label{fig:complete and incomplete image streams}%
\end{figure}

The first limitation is that they assume imaging-based degradation data (including historical data for model training and real-time data for model test) are complete, which means images from all the assets should be collected continuously and regularly with the same sampling time interval (see Figure \ref{fig:complete and incomplete image streams} (a) for an example). In reality, however, engineering assets often operate in harsh environments that significantly impact the quality of collected data due to errors in data acquisition, communication, read/write operations, etc. As a result, degradation images often contain significant levels of missing observations, which is known as incomplete/missing imaging data (see Figure \ref{fig:complete and incomplete image streams} (b) for an example). Such data incompleteness poses a significant challenge for the parameter estimation of existing statistical learning-based prognostic models.

The second common limitation for the existing statistical learning-based prognostic models for applications with imaging data is that they employ unsupervised dimension reduction methods for feature extraction, so there is no guarantee that the extracted features are effective for the subsequent TTF prediction. Specifically, they first use unsupervised dimension reduction methods such as FPCA, PCA, and B Spline to extract features, which are then used to construct prognostic models. Since feature extraction and prognostic model construction are two sequential steps, and no TTF information gets involved in the feature extraction process, it is possible that the extracted features may not be most suitable for predicting TTFs. 

To address the aforementioned challenges, this article proposes a supervised dimension reduction-based prognostic model that uses an asset's incomplete degradation images to predict its TTF. Similar to the existing statistical learning-based prognostic models, the proposed model also consists of two steps: feature extraction and prognostic model construction. However, {unlike the existing models, feature extraction in this article is achieved by developing a new supervised tensor dimension reduction method, which uses historical TTFs to supervise the feature extraction process such that the extracted features are more effective for the subsequent TTF prediction}. In addition, {unlike the existing unsupervised dimension reduction methods that only work for complete imaging data, the proposed supervised dimension reduction method works for both complete and incomplete degradation image streams}. 


The proposed supervised dimension reduction method works as follows: First, it detects a low-dimensional tensor subspace in which the high-dimensional degradation image streams are embedded. This is achieved by constructing an optimization criterion that comprises a feature extraction term and a regression term. The first term extracts low-dimensional features from complete/incomplete degradation image streams of training assets, and the second term builds the connection between these assets' TTFs and the extracted features using LLS regression. LLS regression has been widely used in reliability engineering and survival analysis. It includes a variety of TTF distributions, such as (log)normal, (log)logistic, smallest extreme value, and Weibull, etc., which cover most of the TTF distributions in reality \citep{doray1994ibnr}. Since historical TTFs are used to supervise the feature extraction process, it is expected that the extracted features are more effective for the subsequent prognostic. Solving the optimization criterion of the proposed supervised tensor dimension reduction method yields a set of tensor basis matrices that span the low-dimensional tensor subspace for dimension reduction. We then expand both the historical degradation images in the training data set and real-time degradation images from an asset operating in the field (i.e., test data) using the set of tensor basis matrices to extract the low-dimensional tensor features of the training assets and the test asset. The TTFs of the training assets are then regressed against their tensor features using LLS regression, and the parameters are estimated using maximum likelihood estimation. After that, the tensor features of the test asset are fed into the LLS regression model, and its TTF distribution is predicted. 

To solve the optimization criterion of the proposed supervised dimension reduction method, we will first transfer the criterion into a block multiconvex problem. Next, we will propose a block updating algorithm, which cyclically optimizes one block parameters while keeping other blocks fixed until convergence. In addition, we will demonstrate that when TTFs follow normal or lognormal distributions, each sub problem of the block updating algorithm has a closed-form solution, no matter the degradation image streams are complete or incomplete.

The rest of this paper is orginized as follows. Section \ref{sec:metho} presents the supervised tensor dimension reduction-based prognostic method. Section \ref{sec:optim} introduces the block updating algorithm and closed-form solutions when TTFs follow normal/lognormal distributions. Sections \ref{sec:numer} and \ref{sec:case} validate the effectiveness of the proposed prognostic model using a simulated dataset and data from a rotating machinery, respectively. Section \ref{sec:concl} concludes.

\section{The Methodology}
\label{sec:metho}
In this section, we will introduce the proposed supervised tensor dimension reduction-based prognostic model for applications with incomplete imaging data. 
In subsection 2.1, we will present some basic tensor notations and definitions. subsection 2.2 introduces the supervised tensor dimension reduction method. In subsection 2.3, we will discuss the construction of prognostic model and how to predict the TTF of an asset operating in the field using its real-time degradation imaging data.

\subsection{Preliminaries}

In this section, we introduce some basic notations and definitions of tensor operations that are used throughout the article. The \textit{order} of a tensor is the number of dimensions, also known as ways or modes. Vectors (1 order tensors) are denoted by lowercase boldface letters, e.g., $\bm{s}$. Matrices (2 order tensors) are denoted by boldface uppercase letters, e.g., $\bm{S}$. Higher-order tensors (order is 3 or larger) are denoted by calligraphic letters, e.g., $\mathcal{S}$. Indices are denoted by lowercase letters whose range is from 1 to the uppercase letter of the index, e.g., $n = 1, 2, \ldots, N $. An $N$th-order tensor is denoted as $\mathcal{S} \in \mathbb{R}^{I_1 \times I_2 \times \cdot \cdot \cdot \times I_N}$, where $I_n$ represents the $n$th mode of $\mathcal{S}$. The $(i_1, i_2, \ldots , i_N)$th entry of $ \mathcal{S} \in \mathbb{R}^{I_1 \times I_2 \times \cdot \cdot \cdot \times I_N} $ is denoted by $s_{i_1,i_2,\ldots,i_n}$. A fiber of $\mathcal{S}$ is a vector defined by fixing every index but one. A matrix column is a mode-1 fiber and a matrix row is a mode-2 fiber. The \textit{vectorization} of $\mathcal{S}$, denoted by \textit{vec}($\mathcal{S}$), stacks all the entries of $\mathcal{S}$ into a column vector. The mode-$n$ \textit{matricization} of a tensor $\mathcal{S} \in \mathbb{R}^{I_1 \times I_2 \times \cdot \cdot \cdot \times I_N }$ is denoted by $\bm{S}_{(n)}$, which arranges the mode-$n$ fibers to be the columns of the resulting matrix. The $n$th mode product of a tensor $\mathcal{S} \in \mathbb{R}^{I_1 \times I_2 \times \cdot \cdot \cdot \times I_N}$ and a matrix $\bm{U}_{n} \in \mathbb{R}^{J_n \times I_n}$, denoted by $\mathcal{S} \times_{n} \bm{U}_{n}$, is a tensor whose entry is $(\mathcal{S} \times_{n} \bm{U}_{n})_{i_{1}, \ldots, i_{n-1}, j_{n}, i_{n+1}, \ldots, i_{N}} = \Sigma_{I_n = 1}^{I_N } s_{i_1,\ldots,i_N} u_{j, i_n}$. The \textit{Kronecker product} of two matrices $\bm{A} \in \mathbb{R}^{m \times n}$ and $\bm{B} \in \mathbb{R}^{p \times q}$ is an $mp \times nq$ block matrix:
\begin{equation*}
\bm{A} \otimes \bm{B} = 
\begin{bmatrix}
\bm{a}_{11}\bm{B} & \ldots & \bm{a}_{1n}\bm{B} \\
\vdots & \ddots & \vdots \\
\bm{a}_{m1}\bm{B} & \ldots & \bm{a}_{mn}\bm{B}
\end{bmatrix}.
\end{equation*}

\noindent If $\bm{A}$ and $\bm{B} $ have the same number of columns $n = q$, then the \textit{Khatri-Rao} product is defined as the $mp \times n$ column-wise \textit{Kronecker} product: $\bm{A} \odot \bm{B} = [\bm{a}_1 \otimes \bm{b}_1 \quad \bm{a}_2 \otimes \bm{b}_2 \quad \cdot \cdot \cdot \quad \bm{a}_n \otimes \bm{b}_n]$. If $\bm{a}$ and $\bm{b}$ are vectors, then $\bm{A} \otimes \bm{B} = \bm{A} \odot \bm{B}$. More details about tensor notations and operators can be found in \cite{kolda2009tensor}.


\subsection{The Supervised Tensor Dimension Reduction Method}\label{sec:sub:supervised}

We assume that there exists a historical data set for model training. The data set consists of the degradation image streams of $M$ failed assets along with their TTFs, which are denoted as $\mathcal{X}_m  \in  \mathbb{R}^{I_1 \times I_2 \times  I_3}$ and $y_m \in  \mathbb{R}$, respectively, where $m = 1, 2, \ldots, M  $. For the convenience of introducing the dimension reduction method, we convert the 3D degradation image streams from all the $M$ assets to a 4D tensor $\mathcal{X} \in \mathbb{R}^{I_1 \times I_2 \times I_3 \times M} $, where the sample size $M$ is the 4th mode. Similarly, we let $\bm{y} = (y_1,\ldots,y_M)^\top\in \mathbb{R}^{M \times 1}$ be the vector containing all the TTFs of the $M$ assets. 

Out of the $I_1 \times I_2 \times I_3 \times M $ entries of $\mathcal{X}$,  we use a subset $\Omega \subseteq \lbrace (i_1, i_2, i_3, m),1\leq i_1\leq I_1, 1\leq i_2\leq I_2,1\leq i_3\leq I_3,1\leq m\leq M\rbrace $ to denote the indices of the missing ones. To model the missing data, we define a projection operator $\mathcal{P}_{\Omega}(\cdot)$ as follows:
\begin{equation}\label{eq:omega}
    \mathcal{P}_{\Omega}(\mathcal{X})_{(i_1,i_2,i_{3},m)} = \begin{cases}
    \mathcal{X}_{(i_1,i_2,i_{3},m)}, & \text{if} ~ (i_1,i_2,i_{3},m) \not\in \Omega, \\
    0,  & \text{if} ~ (i_1,i_2,i_{3},m) \in \Omega, 
    \end{cases}
\end{equation}

\noindent where $\mathcal{X}_{(i_1,i_2,i_{3},m)}$ is the $(i_1,i_2,i_{3},m)$-th entry of the 4D tensor $\mathcal{X} \in \mathbb{R}^{I_1 \times I_2 \times I_3 \times M}$. To recover the missing entries in tensor $\mathcal{X}$, we may employ the following Tucker decomposition-based tensor completion method \citep{filipovic2015tucker,liu2012tensor,xu2013parallel}:
\begin{equation}\label{eq:unsupervised}
\min_{\mathcal{S},\bm{U}_{1},\bm{U}_{2},\bm{U}_{3}}~~ ||\mathcal{P}_{\Omega}(\mathcal{X} - \mathcal{S} \times_{1} \bm{U}_{1}^\top \times_{2} \bm{U}_{2}^\top \times_{3} \bm{U}_{3}^\top) ||_{F}^2.  
\end{equation}

\noindent where $|| \cdot ||_{F}^{2}$ is the Frobenius norm, $\bm{U}_{1}\in \mathbb{R}^{P_1\times I_1}$, $\bm{U}_{2}\in \mathbb{R}^{P_2\times I_2}$, $\bm{U}_{3}\in \mathbb{R}^{P_3\times I_3}$ are three factor matrices, $\mathcal{S} \in \mathbb{R}^{P_1 \times P_2 \times P_3 \times M}$ is the low-dimensional core tensor, and $\times_{n}$ is the n-mode product of a tensor with a matrix. The tensor completion criterion \eqref{eq:unsupervised} can be seen as an \textit{unsupervised dimension reduction method} for tensor data with missing entries. This is because the degradation image tensor $\mathcal{X} \in \mathbb{R}^{I_1 \times I_2 \times I_3 \times M} $ is a 4-order tensor, which resides in the tensor (multilinear) space $\mathbb{R}^{I_1} \otimes \mathbb{R}^{I_2} \otimes \mathbb{R}^{I_3} \otimes \mathbb{R}^{M}$, where $\mathbb{R}^{I_1}, \mathbb{R}^{I_2}, \mathbb{R}^{I_3}, \mathbb{R}^{M}$ are the 4 vector (linear) spaces; $\mathcal{S} \in \mathbb{R}^{P_1 \times P_2 \times P_3 \times M}$ can be seen as a feature tensor that resides in the tensor space $\mathbb{R}^{P_1} \otimes \mathbb{R}^{P_2} \otimes \mathbb{R}^{P_3} \otimes \mathbb{R}^{M}$. Usually, we have $P_1\ll I_1$, $P_2\ll I_2$, and $P_3 \ll I_3$ for degradation imaging data due to the high spatio-temporal correlation among pixels. This implies that the dimension of the image stream from the $m$th asset is reduced from $\mathbb{R}^{I_1 \times I_2 \times I_3} $ to $\mathbb{R}^{P_1 \times P_2 \times P_3 }$, where $m=1,\ldots,M$. 

Although criterion \eqref{eq:unsupervised} can be seen as a dimension reduction method, there is no guarantee that the extracted low-dimensional feature tensor $\mathcal{S}$ is effective for the subsequent TTF prediction. To address this challenge, we propose the following \textit{supervised dimension reduction} method by combining a tensor completion term and an LLS regression term:
\begin{equation}\label{eq:not multiconvex}
\min_{\bm{U}_{1},\bm{U}_{2},\bm{U}_{3},\sigma, \bm{\beta}_1,\beta_0,\mathcal{S}}\alpha ||\mathcal{P}_{\Omega}(\mathcal{X} - \mathcal{S} \times_{1} \bm{U}_{1}^\top \times_{2} \bm{U}_{2}^\top \times_{3} \bm{U}_{3}^\top) ||_{F}^2 + (1 - \alpha)\ell(\frac{\bm{y} - \bm{1}_M \beta_{0} - \bm{S}_{(4)}\bm{\beta}_{1}}{\sigma}),
\end{equation}

\noindent where $\bm{y} \in \mathbb{R}^{M \times 1 }$ is the vector containing all the TTFs of the $M$ assets in the training data set. The matrix $\bm{S}_{(4)} \in \mathbb{R}^{M \times (P_1 \times P_2 \times P_3)}$ is the mode-4 matricization of the low-dimensional feature tensor $\mathcal{S}$, the $m$th row of which represents the vectorization of the $m$th asset's feature tensor, $m=1,\ldots,M$. $\beta_0$ is the intercept, and $\bm{\beta_1} \in \mathbb{R}^{(P_1 \times P_2 \times P_3) \times 1}$ is the regression coefficient vector. $\bm{1}_m \in \mathbb{R}^{M \times 1}$ is an $M \times 1$ vector whose entries are all ones. $ \ell(\cdot)$ is the negative log-likelihood function of a location-scale distribution. For example, if TTFs follow normal distributions, then $\ell(\frac{\bm{y} - \bm{1}_M \beta_{0} - \bm{S}_{(4)}\bm{\beta}_{1}}{\sigma}) = \frac{M}{2}\log2\pi + M\log\sigma + \frac{1}{2} \Sigma_{m = 1}^{M} \omega_{m}^{2}$, where $\omega_{m} = \frac{y_m - \beta_{0} - \bm{s}_{(4)}^{m}\bm{\beta}_{1}}{\sigma}$, where $\bm{s}_{(4)}^{m}$ is the $m$th row of $\bm{S}_{(4)}$, and $y_m$ is the TTF of asset $m$; if TTFs follow logistic distributions, then $\ell(\frac{\bm{y} - \bm{1}_M \beta_{0} - \bm{S}_{(4)}\bm{\beta}_{1}}{\sigma}) = M\log\sigma - \Sigma_{m = 1}^{M}\omega_{m} + 2\Sigma_{m = 1}^{M} \log(1 + \exp(\omega_{m}))$; if TTFs follow small extreme value (SEV) distributions, then $\ell(\frac{\bm{y} - \bm{1}_M \beta_{0} - \bm{S}_{(4)}\bm{\beta}_{1}}{\sigma}) = n\log\sigma - \Sigma_{m = 1}^{M}\omega_{m} + \Sigma_{m = 1}^{M}\exp(\omega_{m})$. For assets whose TTFs follow log-location-scale distributions, we may transfer them to the corresponding location-scale distributions by taking their logarithm such that criterion \eqref{eq:not multiconvex} can still be used. For example, log-normal, log-logistics, and Weibull distributions can be transferred to normal, logistics, and SEV distributions, respectively. $\alpha \in [0,1]$ is a weight and $||\cdot||_{F}^{2} $ is the Frobenius norm.

In criterion \eqref{eq:not multiconvex}, the first term $||\mathcal{P}_{\Omega}(\mathcal{X} - \mathcal{S} \times_{1} \bm{U}_{1}^\top \times_{2} \bm{U}_{2}^\top \times_{3} \bm{U}_{3}^\top) ||_{F}^2$ is tensor completion from \eqref{eq:unsupervised}, which reduces the dimension of high-dimensional incomplete degradation image streams and extracts low-dimensional tensor features. The second term $\ell(\frac{\bm{y} - \bm{1}_M \beta_{0} - \bm{S}_{(4)}\bm{\beta}_{1}}{\sigma})$ is LLS regression, which regresses each asset's TTF against its tensor features extracted by the first term. By jointly optimizing the two terms, it is expected that the extracted features are effective for TTF prediction. However, it is challenging to solve criterion (\ref{eq:not multiconvex}) since it is neither convex nor block multi-convex. An optimization problem is block multi-convex when its feasible set and objective function are generally non-convex but convex in each block of variables \citep{xu2013block}. Thus, to simplify the development of optimization algorithms for model parameter estimation, we first transform criterion (\ref{eq:not multiconvex}) to a block multi-convex one. Specifically, we apply the following re-parameterization: $\tilde{\sigma} = 1/\sigma, \tilde{\beta_{0}} = \beta_{0} / \sigma, \tilde{\bm{\beta}_{1}} = \bm{\beta}_{1}/\sigma$. As a result, criterion (\ref{eq:not multiconvex}) can be re-expressed as follows: 
\begin{equation}\label{eq:supervised}
\min_{\bm{U}_{1},\bm{U}_{2},\bm{U}_{3}, \tilde{\sigma}, \tilde{\bm{\beta_1}},\tilde{\beta_0},\bm{S}_{(4)}}\alpha ||\mathcal{P}_{\Omega}(\mathcal{X} - \mathcal{S} \times_{1} \bm{U}_{1}^\top \times_{2} \bm{U}_{2}^\top \times_{3} \bm{U}_{3}^\top) ||_{F}^2 + (1 - \alpha)\ell(\tilde{\sigma}\bm{y} - \bm{1}_M \tilde{\beta_{0}} - \bm{S}_{(4)}\tilde{\bm{\beta}_{1}}),
\end{equation}

\noindent where, $\ell(\tilde{\sigma}\bm{y} - \bm{1}_M \tilde{\beta_{0}} - \bm{S}_{(4)}\tilde{\bm{\beta}_{1}}) = \frac{M}{2}\log2\pi - M\log\tilde{\sigma} + \frac{1}{2} \Sigma_{m = 1}^{M} \tilde{\omega}_{m}^{2}$ for TTFs following normal distributions, and $\tilde{\omega}_{m} = \tilde{\sigma}y_m - \tilde{\beta}_{0} - {\bm{s}}_{(4)}^{m}\tilde{\bm{\beta}}_{1}$, where $\bm{s}_{(4)}^{m}$ is the $m$th row of $\bm{S}_{(4)}$ and $y_m$ is the TTF of asset $m$; $\ell(\tilde{\sigma}\bm{y} - \bm{1}_M \tilde{\beta_{0}} - \bm{S}_{(4)}\tilde{\bm{\beta}_{1}}) = -M\log\tilde{\sigma} - \Sigma_{m = 1}^{M}\tilde{\omega}_{m} + 2\Sigma_{m = 1}^{M} \log(1 + \exp(\tilde{\omega}_{m}))$ for TTFs following logistics distributions, and $\ell(\tilde{\sigma}\bm{y} - \bm{1}_M \tilde{\beta_{0}} - \bm{S}_{(4)}\tilde{\bm{\beta}_{1}}) = -n\log\tilde{\sigma} - \Sigma_{m = 1}^{M}\tilde{\omega}_{m} + \Sigma_{m = 1}^{M}\exp(\tilde{\omega}_{m})$ for TTFs following SEV distributions. 

The optimization algorithm to solve criterion \eqref{eq:supervised}, the value of the weight $\alpha$, and the dimension of the low-dimensional tensor subspace $\{P_1, P_2, P_3\}$ will be discussed in Sections \ref{sec:optim} and \ref{sec:closed}. Solving the optimization criterion \eqref{eq:supervised} using historical training data yields a set of basis matrices $\hat{\bm{U}}_{1}\in\mathbb{R}^{P_1\times I_1}, \hat{\bm{U}}_{2}\in\mathbb{R}^{P_2\times I_2}, \hat{\bm{U}}_{3}\in\mathbb{R}^{P_3\times I_3}$, which contains $P_1$ basis vectors of the $1$-mode linear space $\mathbb{R}^{I_1}$, $P_2$ basis vectors of the $2$-mode linear space $\mathbb{R}^{I_2}$, and $P_3$ basis vectors of the $3$-mode linear space $\mathbb{R}^{I_3}$, respectively. The three linear subspaces form the low-dimensional tensor subspace $\mathbb{R}^{P_1}\otimes \mathbb{R}^{P_2} \otimes \mathbb{R}^{P_3}$ detected by the proposed supervised dimension reduction method. 

One of the assumptions of the proposed supervised tensor dimension reduction method in criterion \eqref{eq:supervised} is that the TTF of the asset follows a distribution from the LLS family. This is reasonable since the LLS family includes a variety of TTF distributions, such as (log)normal, (log)logistic, smallest extreme value, and Weibull, etc., which cover most of the TTF distributions in engineering applications \citep{doray1994ibnr}. Another assumption is that there is a linear relationship between the location parameter and predictors (i.e., degradation signals or their features in this article). Specifically, the location parameters in criterion \eqref{eq:not multiconvex} are expressed as $\bm{1}_M \beta_{0} + \bm{S}_{(4)}\bm{\beta}_{1}$, which are linear weighted combinations of the rows of $\bm{S}_{(4)}$). This assumption is widely used in LLS regression \citep{doray1994ibnr,fang2019image}. However, if a simple linear weighted combination is not adequate to characterize the association between the location parameter and degradation signal features, a high-order polynomial relationship can be constructed \citep{hastie2009elements}. By doing so, we can model a more complex association between the location parameter and features. More importantly, the incorporation of high-order polynomial terms into the proposed supervised tensor dimension reduction method does not affect the effectiveness of the optimization algorithms for parameter estimation to be discussed in Sections \ref{sec:optim} and \ref{sec:closed}.




\subsection{Prognostic Model Construction and Real-Time TTF Prediction}\label{sec:sub:prog}

In this subsection, we discuss how to build a prognostic model based on the supervised dimension reduction method proposed in Section \ref{sec:sub:supervised}, and how to predict the TTF distribution of an asset operating in the field using its real-time degradation image data.

Similar to Section \ref{sec:sub:supervised}, we denote the training data set as $\{\mathcal{X}_m\in\mathbb{R}^{I_1\times I_2\times D_m}, y_m\}_{m=1}^M$, where $M$ is the number of failed assets in the training data set. Notice that $D_m$ might not be the same as $D_{m'}$ for two assets $m$ and $m'$, $m=1,\ldots, M, m'=1,\ldots, M, m\neq m'$. This is because different asset's failure times (i.e., TTFs) are different, and usually no image data can be collected beyond an asset's failure time since the asset is stopped for maintenance or replace once it is failed. In addition to the training data, we denote the degradation image stream of a test asset by time $t$ as $\mathcal{X}_t\in\mathbb{R}^{I_1\times I_2\times I_t}$. The objectives of this subsection include 1) constructing a prognostic model and estimating its parameters using $\{\mathcal{X}_m, y_m\}_{m=1}^M$ in the training data set, and 2) using the estimated prognostic model to predict the TTF (denoted as $\hat y_t$) of the test asset based on its degradation image stream $\mathcal{X}_t$.

We first use the proposed supervised dimension reduction method to extract low-dimensional features of both the training and test assets. Specifically, as discussed in Section \ref{sec:sub:supervised}, we first construct a 4D tensor $\mathcal{X} \in \mathbb{R}^{I_1 \times I_2 \times I_3 \times M} $ using the degradation image streams of the training assets, where $I_3=\max(\{D_m\}_{m=1}^M)$. Note that $\mathcal{X}$ is an incomplete tensor no matter the image streams from the training assets are complete or incomplete. This is because the TTFs of training assets are different, and thus not all the training assets have $I_3$ images. To detect the low-dimensional tensor subspace in which the high-dimensional degradation images are embedded, we solve optimization criterion \eqref{eq:supervised} by using training data $\{\mathcal{X}, \bm{y}\}$, where $\bm{y}=(y_1,\ldots, y_M)^\top$. This yields basis matrices $\hat{\bm{U}}_{1}\in\mathbb{R}^{P_1\times I_1}, \hat{\bm{U}}_{2}\in\mathbb{R}^{P_2\times I_2}, \hat{\bm{U}}_{3}\in\mathbb{R}^{P_3\times I_3}$, which form the low-dimensional tensor subspace $\mathbb{R}^{P_1}\otimes \mathbb{R}^{P_2} \otimes \mathbb{R}^{P_3}$. To extract the low-dimensional features of the training and test assets, we expand the image streams in the low-dimensional tensor subspace $\mathbb{R}^{P_1}\otimes \mathbb{R}^{P_2} \otimes \mathbb{R}^{P_3}$ using the basis matrices $\hat{\bm{U}}_{1},\hat{\bm{U}}_{2},\hat{\bm{U}}_{3}$. This is achieved by solving the following optimization criteria:
\begin{equation}\label{eq:Sm}
\hat{\mathcal{S}}_{m}=\arg \min_{\bm{S}_{m}} ||\mathcal{P}_{\Omega}(\mathcal{X}_{m} - \mathcal{S}_{m} \times_{1} \bm{U}_{1}^\top \times_{2} \bm{U}_{2}^\top \times_{3} \bm{U}_{3}^\top) ||_{F}^2.
\end{equation}
\begin{equation}\label{eq:Sm2}
\hat{\mathcal{S}}_{t}=\arg \min_{\bm{S}_{t}} ||\mathcal{P}_{\Omega}(\mathcal{X}_{t} - \mathcal{S}_{t} \times_{1} \bm{U}_{1}^\top \times_{2} \bm{U}_{2}^\top \times_{3} \bm{U}_{3}^\top) ||_{F}^2,
\end{equation}

\noindent where $\{\hat{\mathcal{S}}_{m}\}_{m=1}^M$ are the low-dimensional feature tensors of the $M$ assets in the training data set, and $\hat{\mathcal{S}}_{t}$ is the low-dimensional feature tensor of the test asset. 

Next, we construct a prognostic model using the low-dimensional feature tensors of the $M$ assets in the training data set (i.e., $\{\hat{\mathcal{S}}_{m}\}_{m=1}^M$) along with their TTFs $\{y_m\}_{m=1}^M$. Specifically, we build the following LLS regression model:
\begin{equation}\label{eq:pred_train2}
    y_{m} = \gamma_{0} + vec(\hat{\mathcal{S}}_{m})^\top \bm{\gamma}_{1} + \sigma\epsilon_{m},
\end{equation}

\noindent where $vec(\hat{\mathcal{S}}_{m})$ is the vectorization of $\hat{\mathcal{S}}_{m}$. $\gamma_{0}\in \mathbb{R}$ and $\bm{\gamma}_{1}\in \mathbb{R}^{(P_1\times P_2\times P_3)\times 1}$ are the regression coefficients, $\sigma$ is the scale parameter, and $\epsilon_{m}$ is the random noise term with a standard location-scale probability density function $f(\epsilon)$. For example, $f(\epsilon) = 1/\sqrt{2\pi}\exp(-\epsilon^{2}/2)$ for a normal distribution and $f(\epsilon) = \exp(\epsilon - \exp(\epsilon))$ for an SEV distribution. The parameters in criterion (\ref{eq:pred_train2}) can be estimated by solving the following optimization problem:
\begin{equation}
    \min_{\bm{y},\gamma_{0}, \bm{\gamma}_{1},\sigma} \ 
\ell(\frac{\bm{y} - \bm{1}_M \gamma_{0} - \hat{\bm{S}}_{(4)}\bm{\gamma}_{1}}{\sigma}), 
\end{equation}

\noindent where  $ \ell(\cdot)$ is the negative log-likelihood function of a location-scale distribution,$\bm{y} = (y_1, y_2, \cdot \cdot \cdot, y_m)^\top$ and $\hat{\bm{S}}_{(4)} = (vec(\hat{\mathcal{S}}_{1})^\top, vec(\hat{\mathcal{S}}_{2})^\top, \cdot \cdot \cdot, vec(\hat{\mathcal{S}}_{M})^\top)^\top$, and $\ell(\cdot)$ is the negative log-likelihood function. We conduct the following reparameterization to transform the optimization to be a convex one: $\tilde{\sigma} = 1/\sigma, \tilde{\gamma}_{0} = \gamma_{0}/ \sigma, \tilde{\bm{\gamma}}_{1} = \bm{\gamma}_{1} / \sigma$:
\begin{equation}\label{eq:pred_train_ga2}
    \{\hat{\tilde{\gamma_{0}}}, \hat{\tilde{\bm{\gamma}}}_{1},\hat{\tilde{\sigma}}\}=\arg \min_{\tilde\gamma_{0}, \tilde{\bm{\gamma}}_{1},\tilde{\sigma}} \ 
\ell(\tilde{\sigma}\bm{y} - \bm{1}_M\tilde\gamma_{0} - \hat{\bm{S}}_{(4)}\tilde{\bm{\gamma}}_{1}).\ 
\end{equation}

\noindent Solving (\ref{eq:pred_train_ga2}) provides the estimated parameters $\{\hat{\tilde{\gamma_{0}}}, \hat{\tilde{\bm{\gamma}}}_{1},\hat{\tilde{\sigma}}\}$, which can be transformed back to the estimation of the parameters in the LLS regression model: $\hat{\gamma}_{0} = \hat{\tilde{\gamma}}_{0} / \hat{\tilde{\sigma}}$, $\hat{\bm{\gamma}}_{1} = \hat{\tilde{\bm{\gamma}}}_{1}/\hat{\tilde{\sigma}}$ and $\hat{\sigma} = 1/ \hat{\tilde{\sigma}}$. As a result, the fitted LLS regression model is $\hat{y}_{m} \sim LLS(\hat{\gamma}_{0} + vec(\hat{\mathcal{S}}_{m})^\top\hat{\bm{\gamma}}_{1}, \hat{\sigma})$, where $\hat{\gamma}_{0} + vec(\hat{\mathcal{S}}_{m})^\top\hat{\bm{\gamma}}_{1}$ and $\hat{\sigma}$ are respectively the estimated location and scale parameters.

Finally, we feed the extracted low-dimensional feature tensor of the test asset into the estimated LLS regression model to predict the asset's TTF distribution:  $\hat{y}_{t} \sim LLS(\hat{\gamma}_{0} + vec(\hat{\mathcal{S}}_{t})^\top\hat{\bm{\gamma}}_{1}, \hat{\sigma})$.

\section{The Optimization Algorithm}
\label{sec:optim}

In this section, we discuss how to solve the supervised tensor dimension reduction method proposed in section \ref{sec:sub:supervised}. In subsection \ref{sec:sub:bua}, we develop a Block Updating Algorithm to solve criterion (\ref{eq:supervised}). The algorithm splits the unknown parameters in criterion \eqref{eq:supervised} into several blocks, and it cyclically optimizes one block parameter while keeping other blocks fixed until convergence. The sub-optimization problem for each block is convex, so the convergence of the block updating algorithm is guaranteed. In subsection \ref{sec:sub:tuning}, we discuss the initialization of the proposed algorithm and hyperparameter tuning.

\subsection{The Block Updating Algorithm}\label{sec:sub:bua}

The Block Updating Algorithm first splits the unknown parameters in criterion \eqref{eq:supervised} into 5 blocks, i.e., $\bm{U}_{1}, \bm{U}_{2}, \bm{U}_{3}, \mathcal{S}$ and $\{\tilde{\beta_{0}}, \tilde{\bm{\beta}_{1}},\tilde{\sigma}\}$. It then cyclically optimizes one block of parameters each time while keeping other blocks fixed.

Specifically, at the $k$th iteration, $\bm{U}_{1}$ is updated by solving the following optimization problem while keeping other blocks (i.e., $\bm{U}_{2}^{k-1}, \bm{U}_{3}^{k-1}, \tilde{\sigma}^{k-1}, \tilde{\beta}_{0}^{k-1}, \tilde{\bm{\beta}}_{1}^{k-1}, \mathcal{S}^{k-1}$) fixed:

\begin{equation}\label{eq:U1k2}
\begin{split}
\bm{U}_{1}^{k} &= \arg\min_{\bm{U}_{1}}~~  \alpha||\mathcal{P}_{\Omega}(\mathcal{X} - \mathcal{S}^{k-1} \times_{1} \bm{U}_{1}^\top \times_{2}  {\bm{U}_{2}^{k-1}}^\top \times_{3} {\bm{U}_{3}^{k-1}}^\top )||_{F}^2 
  +(1 - \alpha)\ell(\tilde{\sigma}^{k-1}, \tilde{\beta}_{0}^{k-1}, \tilde{\bm{\beta}}_{1}^{k-1},\bm{S}_{(4)}^{k-1}) \\& = \arg\min_{\bm{U}_{1}}~~  ||\mathcal{P}_{\Omega}(\mathcal{X} - \mathcal{S}^{k-1} \times_{1} \bm{U}_{1}^\top \times_{2}  {\bm{U}_{2}^{k-1}}^\top \times_{3} {\bm{U}_{3}^{k-1}}^\top )||_{F}^2
\end{split}
\end{equation}
\noindent Similarly, the remaining blocks are updated as follows:
\begin{equation}\label{eq:U2k2}
\begin{split}
\bm{U}_{2}^{k} &= \arg\min_{\bm{U}_{2}}~~ \alpha||\mathcal{P}_{\Omega}(\mathcal{X} - {\mathcal{S}}^{k-1} \times_{1} {\bm{U}_{1}^{k}}^\top \times_{2} {\bm{U}_{2}^\top} \times_{3} {\bm{U}_{3}^{k-1}}^\top) ||_{F}^2 + (1 - \alpha)\ell(\tilde{\sigma}^{k-1}, \tilde{\beta}_{0}^{k-1}, \tilde{\bm{\beta}}_{1}^{k-1},\bm{S}_{(4)}^{k-1}) \\& = \arg\min_{\bm{U}_{2}}~~ ||\mathcal{P}_{\Omega}(\mathcal{X} - {\mathcal{S}}^{k-1} \times_{1} {\bm{U}_{1}^{k}}^\top \times_{2} {\bm{U}_{2}^\top} \times_{3} {\bm{U}_{3}^{k-1}}^\top) ||_{F}^2
\end{split}
\end{equation}
\begin{equation}\label{eq:U3k2}
\begin{split}
\bm{U}_{3}^{k} &= \arg\min_{\bm{U}_{3}}~~\alpha||\mathcal{P}_{\Omega}(\mathcal{X} - {\mathcal{S}}^{k-1} \times_{1} {\bm{U}_{1}^{k}}^\top \times_{2} {\bm{U}_{2}^{k}}^\top \times_{3} {\bm{U}_{3}^\top}) ||_{F}^2  + (1 - \alpha)\ell(\tilde{\sigma}^{k-1}, \tilde{\beta}_{0}^{k-1}, \tilde{\bm{\beta}}_{1}^{k-1},\bm{S}_{(4)}^{k-1}) \\& = \arg\min_{\bm{U}_{3}}~~||\mathcal{P}_{\Omega}(\mathcal{X} - {\mathcal{S}}^{k-1} \times_{1} {\bm{U}_{1}^{k}}^\top \times_{2} {\bm{U}_{2}^{k}}^\top \times_{3} {\bm{U}_{3}^\top}) ||_{F}^2
\end{split}
\end{equation}
\begin{equation}\label{eq:Betak2}
\begin{split}
\{\tilde{\sigma}^{k}, \tilde{\beta}_{0}^{k}, \tilde{\bm{\beta}}_{1}^{k}\} &= \arg\min_{\tilde{\sigma}^{k}, \tilde{\beta}_{0}, \tilde{\bm{\beta}}_{1}}~~\alpha||\mathcal{P}_{\Omega}(\mathcal{X} - {\mathcal{S}}^{k-1} \times_{1} {\bm{U}_{1}^{k}}^\top \times_{2} {\bm{U}_{2}^{k}}^\top \times_{3} {\bm{U}_{3}^{k}}^\top )||_{F}^2 + (1 - \alpha)\ell(\tilde{\sigma},  \tilde{\beta}_{0},  \tilde{\bm{\beta}}_{1}, \bm{S}_{(4)}^{k-1}) \\ &= \arg\min_{\tilde{\sigma}^{k}, \tilde{\beta}_{0}, \tilde{\bm{\beta}}_{1}}~~ \ell(\tilde{\sigma},  \tilde{\beta}_{0},  \tilde{\bm{\beta}}_{1}, \bm{S}_{(4)}^{k-1})
\end{split}
\end{equation}
\begin{equation}\label{eq:Sk2}
\begin{split}
{\mathcal{S}}^{k} = \arg\min_{\mathcal{S}}~~\alpha||\mathcal{P}_{\Omega}(\mathcal{X} - \mathcal{S} \times_{1} {\bm{U}_{1}^{k}}^\top \times_{2} {\bm{U}_2^{k}}^\top &\times_{3} {\bm{U}_{3}^{k}}^\top )||_{F}^2 + (1 - \alpha)\ell(\tilde{\sigma}^{k}, \tilde{\beta}_0^{k}, \tilde{\bm{\beta}}_{1}^{k},\bm{S}_{(4)})
\end{split}
\end{equation}

We summarize the Block Updating Algorithm in Algorithm 1 below. The convergence criterion can be set as  $\Psi(\bm{U}_{1}^{k}, \bm{U}_{2}^{k}, \bm{U}_{3}^{k}, \tilde{\sigma}^{k}, \tilde{\beta}_{0}^{k}, \tilde{\bm{\beta}}_{1}^{k}, \mathcal{S}^{k}) - \Psi(\bm{U}_{1}^{k+1}, \bm{U}_{2}^{k+1}, \bm{U}_{3}^{k+1}, \tilde{\sigma}^{k+1}, \tilde{\beta}_{0}^{k+1}, \tilde{\bm{\beta}}_{1}^{k+1}, \\ \mathcal{S}^{k+1}) < \epsilon $, where $\Psi$ is the value of the objective function in criterion (\ref{eq:supervised}), and $\epsilon$ is a small number. It is easy to show that sub problems \eqref{eq:U1k2}, \eqref{eq:U2k2}, and \eqref{eq:U3k2} are convex. For normal, logistic, and SEV distributions, their negative log-likelihood functions $\ell(\cdot)$ are also convex, so objective functions \eqref{eq:Betak2} and \eqref{eq:Sk2} are convex as well. As a result, the Block Updating Algorithm converges to a stationary point of criterion \eqref{eq:supervised}.
\begin{algorithm}[!htb]\label{alg: algorithm1}
\caption{Block Updating Algorithm for solving criterion \eqref{eq:supervised}}

\textbf{Input:} Tensor $\mathcal{X}$ constructed from the (incomplete) degradation image streams of
$M$ assets and the TTF vector $\bm{y}$; the dimension of the low-dimensional 
tensor subspace $\{P_1,P_2,P_3\} $  

\textbf{Initialization:} Initialize $(\bm{U}_{1}^{0}, \bm{U}_{2}^{0}, \bm{U}_{3}^{0}, \tilde{\sigma}^{0}, \tilde{\beta}_{0}^{0}, \tilde{\bm{\beta}}_{1}^{0},\mathcal{S}^{0})$ randomly or heuristically

\textbf{While} convergence criterion not met \textbf{do}

$\bm{U}_{1}^{k} \longleftarrow $ (\ref{eq:U1k2}) 

$\bm{U}_{2}^{k} \longleftarrow$  (\ref{eq:U2k2}) 

$\bm{U}_{3}^{k} \longleftarrow$  (\ref{eq:U3k2}) 

$(\tilde{\sigma}^{k}, \tilde{\beta}_{0}^{k}, \tilde{\bm{\beta}}_{1}^{k}) \longleftarrow$ (\ref{eq:Betak2}) 

$\mathcal{S}^{k} \longleftarrow$ (\ref{eq:Sk2})

$k = k + 1 $
\textbf{End While}

\textbf{Output:} Basis matrices of the low-dimensional tensor subspace $\{\bm{U}_{1}^{k}, \bm{U}_{2}^{k}, \bm{U}_{3}^{k}\}$
\end{algorithm}

\subsection{Initialization and Hyperparameter Tuning}\label{sec:sub:tuning}

To run Algorithm 1, we need to initialize the parameters $\bm{U}_{1}^{0}, \bm{U}_{2}^{0}, \bm{U}_{3}^{0}, \tilde{\sigma}^{0}, \tilde{\beta}_{0}^{0}, \tilde{\bm{\beta}}_{1}^{0},\mathcal{S}^{0}$. The initialization can be accomplished randomly or heuristically. In this article, we propose a heuristic initialization method. Specifically, if tensor $\mathcal{X}$ has no missing entries, MPCA \citep{lu2008mpca} is applied to tensor $\mathcal{X}$, which yields $\{\bm{U}_{1}^{0}, \bm{U}_{2}^{0}, \bm{U}_{3}^{0}\}$. Next, we compute ${\mathcal{S}}^{0}$ by solving ${\mathcal{S}}^{0} = \arg\min_{\mathcal{S}}~~||\mathcal{P}_{\Omega}(\mathcal{X} - \mathcal{S} \times_{1} {\bm{U}_{1}^{0}}^\top \times_{2} {\bm{U}_{2}^{0}}^\top \times_{3} {\bm{U}_{3}^{0}}^\top) ||_{F}^2$. Finally, $ \tilde{\beta}_{0}^{0}, \tilde{\bm{\beta}}_{1}^{0}, \tilde{\sigma}^{0}$ are computed by solving $    \min_{\tilde{\beta}_{0}^{0}, \tilde{\bm{\beta}}_{1}^{0},\tilde{\sigma}^{0}} \ 
\ell(\tilde{\sigma}^{0}\bm{y} - \bm{1}_M \tilde{\beta}_{0}^{0} - \bm{S}_{(4)}^{0}\tilde{\bm{\beta}}_{1}^{0})$, where $\bm{S}_{(4)}^{0}$ is the mode-4 matricization of $\mathcal{S}^{0}$. If tensor $\mathcal{X}$ has missing values, a tensor completion method \citep{filipovic2015tucker,liu2012tensor,xu2013parallel} can be conducted before applying MPCA.

In addition to the initialization, the hyperparameter parameters including the weight $\alpha$ and the dimension of tensor subspace $(P_1, P_2, P_3)$ also need to be pre-determined. It is known that $\alpha$ controls the weights of the feature extraction term and the regression term, and $\alpha\in[0,1]$. To select an appropriate weight parameter, we will first split the range $[0,1]$ into $L+1$ intervals equally, which yields $\alpha_0=0/L,\alpha_1=1/L,\alpha_2=2/L,\ldots,\alpha_L=L/L$. Next, we employ cross-validation to select the weight that achieves the highest prediction accuracy. If the weight at the boundary is selected  (i.e., $\alpha_0=0/L$ or $\alpha_L=L/L$), we further split the interval closest to the boundary and conduct cross-validation again. For example, if $\alpha_0=0/L$ is chosen as the best weight, we will split $[0/L,1/L]$ into $(L+1)$ intervals equally and re-conduct the cross-validation. This process is repeated until a non-boundary weight is selected. Of course, a maximum number of repetitions needs to be set to control the computational time. 

The values of $\{P_1,P_2,P_3\}$ can be determined using cross-validation as well. To be specific, we may try a certain number of candidate values for $\{P_1,P_2,P_3\}$ and run Algorithm 1 to extract low-dimensional features, which are then used to build the prognostic model discussed in Section \ref{sec:sub:prog} for TTF prediction. The values which achieve the smallest prediction error will be chosen. It is known that there usually exists high spatio-temporal correlations among degradation image streams \citep{fang2019image}, so the dimension of the tensor subspace is usually low, which helps reduce the computation intensity of model selection. The values of $\{P_1,P_2,P_3\}$ can also be determined heuristically. For example, if MPCA is employed for parameter initialization, then the fraction-of-variance-explained \citep{lu2008mpca} can be used to determine the dimension of tensor subspace.

\section{Analytical Solutions}\label{sec:closed}

In this section, we discuss the closed-form solutions of optimization problems \eqref{eq:U1k2}, \eqref{eq:U2k2}, \eqref{eq:U3k2}, \eqref{eq:Betak2}, and \eqref{eq:Sk2} in Algorithm 1. Specifically, we will discuss the solutions when degradation image streams are complete and incomplete in Sections \ref{sec:sub:llscomplete} and \ref{sec:sub:llsincomplete}, respectively. For simplicity, we will remove the superscripts $k$ and $k-1$.

\subsection{Analytical Solutions for Complete Data}\label{sec:sub:llscomplete}

\subsubsection{Solution procedure for $\bm{U}_{1}$}

When degradation image streams are complete (i.e., the 4D image tensor $\mathcal{X}$ in criterion \eqref{eq:supervised} has no missing entries), we have the following proposition, which provides the analytical solution to problem \eqref{eq:U1k2}.

\begin{prop}
If the 4D tensor $\mathcal{X}$ has no missing values, optimization problem (\ref{eq:U1k2}) has the following analytical solution
\begin{equation*}
    \bm{U}_{1} = (\bm{X}_{(1)} \cdot \bm{S}_{{U}_{1}(1)}^\top \cdot ( \bm{S}_{{U}_{1}(1)} \cdot \bm{S}_{{U}_{1}(1)}^\top)^{-1})^\top,
\end{equation*}

\noindent where $\bm{X}_{(1)}$ is the mode-1 matricization of $\mathcal{X}$, $\mathcal{S}_{{U}_{1}}=\mathcal{S} \times_{2} {\bm{U}}_{2}^\top \times_{3} {\bm{U}}_{3}^\top$, $\bm{S}_{{U}_{1}(1)}$ is the mode-1 matricization of $\mathcal{S}_{{U}_{1}}$, and the operator ``$\cdot$" represents multiplication.
\end{prop}

\subsubsection{Solution procedure for $\bm{U}_{2}$}

When degradation image streams are complete, the proposition below gives the analytical solution to problem \eqref{eq:U2k2}.

\begin{prop}
If the 4D tensor $\mathcal{X}$ has no missing values, optimization problem  (\ref{eq:U2k2}) has the following analytical solution
\begin{equation*}
    \bm{U}_{2} = (\bm{X}_{(2)} \cdot \bm{S}_{{U}_{2}(2)}^\top \cdot ( \bm{S}_{{U}_{2}(2)} \cdot \bm{S}_{{U}_{2}(2)}^\top)^{-1})^\top,
\end{equation*}

\noindent where $\bm{X}_{(2)}$ is the mode-2 matricization of $\mathcal{X}$, $\mathcal{S}_{{U}_{2}}=\mathcal{S} \times_{1} {\bm{U}}_{1}^\top \times_{3} {\bm{U}}_{3}^\top$, and $\bm{S}_{{U}_{2}(2)}$ is the mode-2 matricization of $\mathcal{S}_{{U}_{2}}$.
\end{prop}

\subsubsection{Solution procedure for $\bm{U}_{3}$}

When degradation image streams are complete, the proposition below provides the analytical solution to problem \eqref{eq:U3k2}.

\begin{prop}
If the 4D tensor $\mathcal{X}$ has no missing values, optimization problem  (\ref{eq:U3k2}) has the following analytical solution
\begin{equation*}
    \bm{U}_{3} = (\bm{X}_{(3)} \cdot \bm{S}_{{U}_{3}(3)}^\top \cdot ( \bm{S}_{{U}_{3}(3)} \cdot \bm{S}_{{U}_{3}(3)}^\top)^{-1})^\top,
\end{equation*}

\noindent where $\bm{X}_{(3)}$ is the mode-3 matricization of $\mathcal{X}$, $\mathcal{S}_{{U}_{3}}=\mathcal{S} \times_{1} {\bm{U}}_{1}^\top \times_{2} {\bm{U}}_{2}^\top$, and $\bm{S}_{{U}_{3}(3)}$ is the mode-3 matricization of $\mathcal{S}_{{U}_{3}}$.
\end{prop}

\subsubsection{Solution procedure for $\tilde{\beta}_{0}, \tilde{\bm{\beta}}_{1},\tilde{\sigma}$}\label{sec:sub:sub:sigma}

For general LLS distributions, there is no closed-form solution for $\tilde{\beta}_{0}, \tilde{\bm{\beta}}_{1},\tilde{\sigma}$. As a result, we may use existing algorithms \citep{doray1994ibnr} or convex optimization packages to solve problem \eqref{eq:Betak2}. However, if the TTF follows a Normal (or lognormal) distribution, we may replace the negative log-likelihood term in criterion \eqref{eq:not multiconvex} with a mean squared error-based loss function, which results in the following optimization criterion:
\begin{equation}\label{eq:normal (lognormal)}
\min_{\bm{U}_{1},\bm{U}_{2},\bm{U}_{3}, \bm{\beta}_1,\beta_0,\mathcal{S}}\alpha ||\mathcal{P}_{\Omega}(\mathcal{X} - \mathcal{S} \times_{1} \bm{U}_{1}^\top \times_{2} \bm{U}_{2}^\top \times_{3} \bm{U}_{3}^\top) ||_{F}^2 + (1 - \alpha)\|\bm{y} - \bm{1}_M \beta_{0} - \bm{S}_{(4)}\bm{\beta}_{1}\|_2^2.
\end{equation}

Since criterion \eqref{eq:normal (lognormal)} is multi-convex, no re-parameterization is needed. As a result, problem \eqref{eq:Betak2} is equivalent to $\min_{\beta_{0}, \bm{\beta}_{1}} \|\bm{y} - \bm{1}_M \beta_0 - \bm{S}_{(4)}\bm{\beta}_{1}\|_2^2$, the solution to which can be easily found using least squares: $\bm{\beta}=(\bm{S}^\top\bm{S})^{-1}\bm{S}^\top\bm{y}$, where $\bm{\beta}=(\beta_0, \bm{\beta}^\top_1)^\top\in\mathbb{R}^{(P_1 \times P_2 \times P_3+1)\times 1}$ and $\bm{S}=(\bm{1}_M, \bm{S}_{(4)})\in\mathbb{R}^{M\times (P_1 \times P_2 \times P_3+1)}$. With the estimation of the regression coefficients, the scale parameter can be estimated using $\sigma=\sqrt{\|\bm{y} - \bm{1}_M \beta_{0} - \bm{S}_{(4)}\bm{\beta}_{1}\|_2^2/(M-\\P_1\times P_2\times P_3 -1)}$. 

\subsubsection{Solution procedure for $\mathcal{S}$}\label{sec:sub:sub:Scomplete}

For general LLS distributions, there is no closed-form solution for $\mathcal{S}$ either. Therefore, we may use existing convex optimization packages to solve problem \eqref{eq:Sk2}. However, if the TTF follows a Normal (or lognormal) distribution, according to criterion \eqref{eq:normal (lognormal)}, problem \eqref{eq:Sk2} is equivalent to 
\begin{equation}\label{eq:Sknormal}
\begin{split}
{\mathcal{S}}^{k} = \arg\min_{\mathcal{S}}~~\alpha||\mathcal{P}_{\Omega}(\mathcal{X} - \mathcal{S} \times_{1} {\bm{U}_{1}^{k}}^\top &\times_{2} {\bm{U}_2^{k}}^\top \times_{3} {\bm{U}_{3}^{k}}^\top )||_{F}^2  + (1 - \alpha)\|\bm{y} - \bm{1}_M \beta_0^{k} - \bm{S}_{(4)}\bm{\beta}_{1}^{k}\|_2^2,
\end{split}
\end{equation}
\noindent which has an analytical solution as suggested by Proposition 4.

\begin{prop}
If the 4D tensor $\mathcal{X}$ has no missing values, optimization problem  (\ref{eq:Sknormal}) has the following analytical solution
\begin{equation*}
\begin{split}
    \bm{S}_{(4)} &= \big[\alpha \cdot \bm{X}_{(4)} \cdot (\bm{U}_{3} \otimes \bm{U}_{2} \otimes \bm{U}_{1})^\top + (1 - \alpha) \cdot (\bm{y} - \bm{1}_{M} \cdot \beta_{0}) \cdot \bm{\beta}_{1}^\top \big] \cdot \\
    &
     \big[\alpha \cdot (\bm{U}_{3} \otimes \bm{U}_{2} \otimes \bm{U}_{1}) \cdot (\bm{U}_{3} \otimes \bm{U}_{2} \otimes \bm{U}_{1})^\top +(1 - \alpha) \cdot \bm{\beta}_{1} \cdot \bm{\beta}_{1}^\top \big]^{-1},
\end{split}
\end{equation*}

\noindent where $\bm{X}_{(4)}$ is the mode-4 matricization of $\mathcal{X}$, $\bm{S}_{(4)}$ is the mode-4 matricization of $\mathcal{S}$.
\end{prop}


The proof of Propositions 1, 2, 3, and 4 can be found in the Appendix

\subsection{Analytical Solutions for Incomplete Data}\label{sec:sub:llsincomplete}

In this subsection, we discuss the closed-form solutions for optimization problems \eqref{eq:U1k2}, \eqref{eq:U2k2}, \eqref{eq:U3k2}, \eqref{eq:Betak2}, and \eqref{eq:Sk2} when the degradation tensor $\mathcal{X}$ in criterion \eqref{eq:supervised} is incomplete. We consider the most general missing pattern, \textit{entry-wise missing}, which means that any entry of $\mathcal{X}\in \mathbb{R}^{I_1 \times I_2 \times I_3 \times M}$ can be missing. Thus, the indices of missing entries can be denoted as a subset $\Omega \subseteq \{ (i_1, i_2, i_3, m),1\leq i_1\leq I_1, 1\leq i_2\leq I_2,1\leq i_3\leq I_3,1\leq m\leq M\}$.


\subsubsection{Solution procedure for $\bm{U}_{1}$}

When tensor $\mathcal{X}$ has a general entry-wise missing structure, there is no closed-form solution for $\bm{U}_{1}$ in optimization criterion \eqref{eq:U1k2}. However, we may decompose criterion \eqref{eq:U1k2} into multiple sub-optimization problems, each of which has an analytical solution. To be specific, we first give the following lemma. 

\begin{lemma}\label{lemma1}
Let $\bm{A} \in \mathbb{R}^{M \times N}$, $\bm{C} \in \mathbb{R}^{P \times N}$, and $\bm{B} \in \mathbb{R}^{M \times P}$, the solution to criterion $\arg\min_{B} \|\bm{A} - \bm{B}\bm{C} \|_{F}^{2}$ can be found by solving each row of $\bm{B}$ separately--that is--solving $\{\bm{b}_{m}\}_{m=1}^{M}$ as follows:
\begin{equation*}
\arg\min_{b_m} \| \bm{a}_{m} - \bm{b}_{m}\bm{C} \|_{F}^{2}, \quad m = 1, \ldots, M
\end{equation*}

\noindent where $\bm{a}_{m}\in \mathbb{R}^{1 \times N}$ is the $m$th row of $\bm{A}$, $\bm{b}_{m} \in \mathbb{R}^{1 \times P} $ is the $m$th row of $\bm{B}$.
\end{lemma}

The proof of Lemma 1 can be found in the Appendix. Lemma 1 enables us to solve each column of matrices $\bm{U}_{1}$ separately. Denote the $i_1$th column of matrix $\bm{U}_{1} \in\mathbb{R}^{P_1 \times I_1}$ as $\bm{u}_{1}^{i_1}\in\mathbb{R}^{P_1 \times 1}, i_1=1,\ldots, I_1$, we replace optimization problem \eqref{eq:U1k2} with $I_1$ sub problems by separately optimizing $\bm{u}_{1}^{1}, \bm{u}_{1}^{2},\ldots, \bm{u}_{1}^{I_1}$. Proposition 5 suggests that there is an analytical solution when optimizing $\bm{u}_{1}^{i_1}$. 
\begin{prop}
When optimizing the $i_1$th column of $\bm{U}_{1}$ in problem \eqref{eq:U1k2}, we have the following analytical solution
\begin{equation*}
\bm{u}_{1}^{i_1} = (\bm{x}_{(1)}^{i_1,{\pi_{i_1}}} \cdot {\bm{S}_{{U}_{1}(1)}^{\pi_{i_1}}}^\top \cdot ( \bm{S}_{{U}_{1}(1)}^{\pi_{i_1}} \cdot {\bm{S}_{{U}_{1}(1)}^{\pi_{i_1}}}^\top)^{-1})^\top,
\end{equation*}

\noindent where $\bm{x}_{(1)}^{i_1}$ denotes the $i_1$th row of $\bm{X}_{(1)}$, $\pi_{i_1}$ is a set consisting of the indices of available entries of $\bm{x}_{(1)}^{i_1}$,  $\bm{x}_{(1)}^{i_1,\pi_{i_1}}$ denotes a vector consisting of the available entries in the $i_1$th row of $\bm{X}_{(1)}$, $\mathcal{S}_{{U}_{1}} = \mathcal{S} \times_{2} \bm{U}_{2}^\top \times_{3} \bm{U}_{3}^\top$, $\bm{S}_{{U}_{1}(1)}$ is the mode-1 matricization of $\mathcal{S}_{{U}_{1}}$, and $\bm{S}_{{U}_{1}(1)}^{\pi_{i_1}} $ denotes a matrix comprises the $\pi_{i_1}$ columns of $\bm{S}_{{U}_{1}(1)} $.
\end{prop}

As mentioned earlier, when tensor $\mathcal{X}$ has a general entry-wise missing structure, there is no closed-form solution for $\bm{U}_{1}$ in optimization criterion \eqref{eq:U1k2}. Therefore, we have to optimize each column of matrix $\bm{U}_{1}$ separately using the analytical solution provided in Proposition 5. However, for image-based applications, the missing data in tensor $\mathcal{X}$ are images but not entries, as illustrated in Figure \ref{fig:degradation stream images}(b). This yields an \textit{image-wise missing} structure, which is a special case of the general \textit{entry-wise missing} structure. For image streams with an \textit{image-wise missing} structure, Proposition 6 suggests that $\bm{U}_{1}$ in optimization criterion \eqref{eq:U1k2} can be estimated analytically as a whole, which means we do not have to optimize each of its columns separately. 

\begin{prop}
If the indices of tensor $\mathcal{X}$'s missing entries can be denoted as $\Omega \subseteq \lbrace (:, :, i_3, m)$, $1\leq i_3\leq I_3,1\leq m\leq M\}$, where ``$:$" denotes all the indices in a dimension, then $\mathcal{X}$'s mode-1 matricization $\bm{X}_{(1)}$ has missing columns. Let $\pi$ be the set consisting of the indices of available columns in $\bm{X}_{(1)}$, then optimization problem (\ref{eq:U1k2}) has the following analytical solution
\begin{equation*}
    \bm{U}_{1} = (\bm{X}_{(1)}^{\pi} \cdot {\bm{S}_{{U}_{1}(1)}^{\pi}}^\top \cdot ( \bm{S}_{{U}_{1}(1)}^{\pi} \cdot {\bm{S}_{{U}_{1}(1)}^{\pi}}^\top)^{-1})^\top,
\end{equation*}

\noindent where $\bm{X}_{(1)}^{\pi}$ is a matrix consisting of the $\pi$ columns of $\bm{X}_{(1)}$, $\mathcal{S}_{{U}_{1}} = \mathcal{S} \times_{2} \bm{U}_{2}^\top \times_{3} \bm{U}_{3}^\top$, $\bm{S}_{{U}_{1}(1)}$ is the mode-1 matricization of $\mathcal{S}_{{U}_{1}}$, and $\bm{S}_{{U}_{1}(1)}^{\pi}$ denotes a matrix constituting the $\pi$ columns of $\bm{S}_{{U}_{1}(1)}$. 
\end{prop}

\subsubsection{Solution procedure for $\bm{U}_{2}$}

Similar to $\bm{U}_{1}$, there is no closed-form solution for $\bm{U}_{2}$ in optimization criterion \eqref{eq:U2k2} when tensor $\mathcal{X}$ has a general entry-wise missing structure. However, Lemma 1 implies that we may also decompose optimization problem \eqref{eq:U2k2} into multiple sub-criteria, each of which has a closed-form solution. Specifically, denote the $i_2$th column of matrix $\bm{U}_{2} \in\mathbb{R}^{P_2 \times I_2}$ as $\bm{u}_{2}^{i_2}\in\mathbb{R}^{P_2 \times 1}, i_2=1,\ldots, I_2$, we can replace optimization problem \eqref{eq:U2k2} with $I_2$ sub problems by separately optimizing $\bm{u}_{2}^{1}, \bm{u}_{2}^{2},\ldots, \bm{u}_{2}^{I_2}$. Proposition 7 shows that there is an analytical solution for $\bm{u}_{2}^{i_2}$.
\begin{prop}
When optimizing the $i_2$th column of $\bm{U}_{2}$ in problem \eqref{eq:U2k2}, we have the following analytical solution
\begin{equation*}
\bm{u}_{2}^{i_2} = (\bm{x}_{(2)}^{i_2,\pi_{i_2}} \cdot {\bm{S}_{{U}_{2}(2)}^{\pi_{i_2}}}^\top \cdot ( \bm{S}_{{U}_{2}(2)}^{\pi_{i_2}} \cdot {\bm{S}_{{U}_{2}(2)}^{\pi_{i_2}}}^\top)^{-1})^\top
\end{equation*}

\noindent where $\bm{x}_{(2)}^{i_2}$ denotes the $i_2$th row of $\bm{X}_{(2)}$, $\pi_{i_2}$ is a set consisting of the indices of available entries of $\bm{x}_{(2)}^{i_2}$, $\bm{x}_{(2)}^{i_2,\pi_{i_2}}$ is a vector consisting of the available entries in the $i_2$th column of $\bm{X}_{(2)}$, $\mathcal{S}_{{U}_{2}} = \mathcal{S} \times_{1} \bm{U}_{1}^\top \times_{3} \bm{U}_{3}^\top$, $\bm{S}_{{U}_{2}(2)}$ is the mode-2 matricization of $\mathcal{S}_{{U}_{2}}$, and $\bm{S}_{{U}_{2}(2)}^{\pi_{i_2}} $ denotes a matrix comprises the $\pi_{i_2}$ columns of $\bm{S}_{{U}_{2}(2)} $.
\end{prop}

Similar to $\bm{U}_{1}$, when tensor $\mathcal{X}$ has the \textit{image-wise missing} structure, we do not have to optimize each of the columns of $\bm{U}_{2}$ separately. Proposition 8 below gives an analytical solution to $\bm{U}_{2}$ when tensor $\mathcal{X}$ has missing images.
\begin{prop}

If the indices of tensor $\mathcal{X}$'s missing entries can be denoted as $\Omega \subseteq \lbrace (:, :, i_3, m)$, $1\leq i_3\leq I_3,1\leq m\leq M\}$, where ``$:$" denotes all the indices in a dimension, then $\mathcal{X}$'s mode-2 matricization $\bm{X}_{(2)}$ has missing columns. Let $\pi$ be the set consisting of the indices of available columns in $\bm{X}_{(2)}$, then optimization problem (\ref{eq:U2k2}) has the following analytical solution
\begin{equation*}
    \bm{U}_{2} = (\bm{X}_{(2)}^{\pi} \cdot {\bm{S}_{{U}_{2}(2)}^{\pi}}^\top \cdot ( \bm{S}_{{U}_{2}(2)}^{\pi} \cdot {\bm{S}_{{U}_{2}(2)}^{\pi}}^\top)^{-1})^\top,
\end{equation*}

\noindent where $\bm{X}_{(2)}^{\pi}$ is a matrix consisting of the $\pi$ columns of $\bm{X}_{(2)}$, $\bm{S}_{{U}_{2}(2)}$ is the mode-2 matricization of $\mathcal{S}_{{U}_{2}}$,  and $\bm{S}_{{U}_{2}(2)}^{\pi}$ denotes a matrix constituting the $\pi$ columns of $\bm{S}_{{U}_{2}(2)}$.  
\end{prop}

\subsubsection{Solution procedure for $\bm{U}_{3}$}

There is no closed-form solution for $\bm{U}_{3}$ in optimization criterion \eqref{eq:U3k2} when tensor $\mathcal{X}$ has missing entries. Based on Lemma 1, we decompose optimization problem \eqref{eq:U3k2} into multiple sub-criteria, each of which has a closed-form solution. Denote the $i_3$th column of matrix $\bm{U}_{3} \in\mathbb{R}^{P_3 \times I_3}$ as $\bm{u}_{3}^{i_3}\in\mathbb{R}^{P_3 \times 1}, i_3=1,\ldots, I_3$, we replace optimization problem \eqref{eq:U3k2} with $I_3$ sub problems by separately optimizing $\bm{u}_{3}^{1}, \bm{u}_{3}^{2},\ldots, \bm{u}_{3}^{I_3}$ respectively. Proposition 9 suggests that there is an analytical solution when optimizing $\bm{u}_{3}^{i_3}$.

\begin{prop}
When optimizing the $i_3$th column of $\bm{U}_{3}$ in problem \eqref{eq:U3k2}, we have the following analytical solution
\begin{equation*}
\bm{u}_{3}^{i_3} = (\bm{x}_{(3)}^{i_3,\pi_{i_3}} \cdot {\bm{S}_{{U}_{3}(3)}^{\pi_{i_3}}}^\top \cdot ( \bm{S}_{{U}_{3}(3)}^{\pi_{i_3}} \cdot {\bm{S}_{{U}_{3}(3)}^{\pi_{i_3}}}^\top)^{-1})^\top,
\end{equation*}

\noindent where $\bm{x}_{(3)}^{i_3}$ denotes the $i_3$th row of $\bm{X}_{(3)}$, $\pi_{i_3}$ is a set consisting of the indices of available entries of $\bm{x}_{(3)}^{i_3}$, $\bm{x}_{(3)}^{i_3,\pi_{i_3}}$ is a vector consisting of the available entries in the $i_3$th row of $\bm{X}_{(3)}$, $\mathcal{S}_{{U}_{3}} = \mathcal{S} \times_{1} \bm{U}_{1}^\top \times_{2} \bm{U}_{2}^\top$, $\bm{S}_{{U}_{3}(3)}$ is the mode-3 matricization of $\mathcal{S}_{{U}_{3}}$, and $\bm{S}_{{U}_{3}(3)}^{\pi_{i_3}} $ denotes a matrix comprises the $\pi_{i_3}$ columns of $\bm{S}_{{U}_{3}(3)} $.
\end{prop}

\subsubsection{Solution procedure for $\tilde{\beta}_{0}, \tilde{\bm{\beta}}_{1},\tilde{\sigma}$}

Whether tensor $\mathcal{X}$ contains missing entries does not affect the methods for $\tilde{\beta}_{0}, \tilde{\bm{\beta}}_{1}$, and $\tilde{\sigma}$ estimation. Therefore, the estimation methods discussed in Section \ref{sec:sub:sub:sigma} can still be used.

\subsubsection{Solution procedure for $\mathcal{S}$}

For general LLS distributions, there is no closed-form solution for $\mathcal{S}$. Therefore, we may use existing convex optimization packages to solve problem \eqref{eq:Sk2}. However, if the TTF follows a Normal (or lognormal) distribution, problem \eqref{eq:Sk2} is equivalent to \eqref{eq:Sknormal} (see Section \ref{sec:sub:sub:Scomplete} for details). Based on Lemma 1, we may optimize each row of $\bm{S}_{(4)}$ separately. We denote the $m$th row of matrix $\bm{S}_{(4)}\in \mathbb{R}^{M \times (P_1 \times P_2 \times P_3)}$ as $\bm{s}_{(4)}^{m}\in \mathbb{R}^{1 \times (P_1 \times P_2 \times P_3)}$, $m=1,\ldots, M$, and replace optimization problem \eqref{eq:Sknormal} with $M$ sub problems--that is--separately optimizing $\bm{s}_{(4)}^{1}, \bm{s}_{(4)}^{2},\ldots, \bm{s}_{(4)}^{M}$. Proposition 10 suggests that there is an analytical solution when optimizing $\bm{s}_{(4)}^{m}$.

\begin{prop}

When optimizing the $m$th row of matrix $\bm{S}_{(4)}$ in problem (\ref{eq:Sknormal}), we have the following analytical solution
\begin{equation*}
\begin{split}
\bm{s}_{(4)}^{m} &= \big[\alpha \cdot \bm{x}_{(4)}^{m,\pi_m} \cdot (\bm{U}_{3} \otimes \bm{U}_{2} \otimes \bm{U}_{1})^{\pi_m \top} + (1 - \alpha) \cdot (y_{m} - \tilde{\beta}_{0}) \cdot \tilde{\bm{\beta}}_{1}^\top \big] \cdot \\
&
\big[\alpha \cdot (\bm{U}_{3} \otimes \bm{U}_{2} \otimes \bm{U}_{1})^{\pi_m} \cdot (\bm{U}_{3} \otimes \bm{U}_{2} \otimes \bm{U}_{1})^{\pi_m \top} + (1 - \alpha) \cdot \tilde{\bm{\beta}}_{1} \cdot \tilde{\bm{\beta}}_{1}^\top\big]^{-1},
\end{split}
\end{equation*}

\noindent where $\bm{x}_{(4)}^{m}$ represents the $m$th row of $\bm{X}_{(4)}$, $\pi_m$ denotes the set consisting of the indices of available entries in $\bm{x}_{(4)}^{m}$,  $\bm{x}_{(4)}^{m,\pi_m}$ is a vector consisting of the available entries in the $m$th row of $\bm{X}_{(4)}$, and $(\bm{U}_{3} \otimes \bm{U}_{2} \otimes \bm{U}_{1})^{\pi_m}$ denotes a matrix comprising the $\pi_m$ columns of matrix $\bm{U}_{3} \otimes \bm{U}_{2} \otimes \bm{U}_{1}$. 
\end{prop}

The proof of all Propositions 5, 6, 7, 8, 9, and 10 can be found in the Appendix.


\color{black}

\section{Numerical Studies}
\label{sec:numer}
In this section, we validate the effectiveness of our proposed {supervised tensor dimension reduction}-based prognostic model using simulated data.

\subsection{Data Generation}
\begin{figure}[htp!]
	\centering
	\subfigure[Without noise]{\raisebox{+0cm}{\includegraphics[width=15cm]{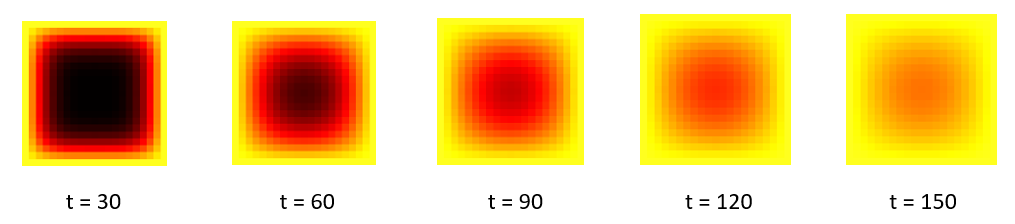} }}%
	\mbox{}
	\subfigure[With noise]{{\includegraphics[width=15cm]{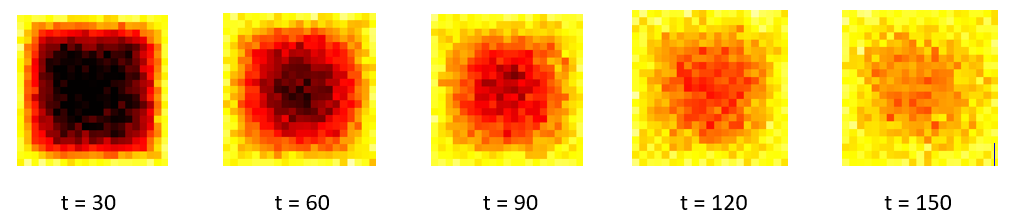} }}%
	\caption{Simulated degradation images based on heat transfer process.} %
	\label{fig:degradation stream images}%
\end{figure}

We generate degradation image streams for $500$ assets. The image stream from asset $m$, which is denoted by $\mathcal{X}_{m}(x,y,t), ~m = 1, 2,  \ldots, 500$, is generated from the following heat transfer equation:
\begin{equation}
\frac{\partial \mathcal{X}_{m}(x,y,t) }{\partial t} = 
\alpha_{m} \Big( \frac{\partial^{2} \mathcal{X}_{m}}{\partial x^{2}} + \frac{\partial^{2} \mathcal{X}_{m}}{\partial y^{2}}\Big),
\end{equation}

\noindent where $ (x, y), 0 \le x, y \le 0.2 $, represents the location of each image pixel. $\alpha_{m}$ is the thermal diffusivity coefficient, which is randomly generated from a uniform distribution $\mathcal{U}(0.5 \times 10^{-4}, 1 \times 10^{-4})$. $t$ is the time index. The initial and boundary conditions are set such that $\mathcal{X}|_{t = 1} = 0 $ and $\mathcal{X}_m|_{x = 0} = \mathcal{X}_m|_{x = 0.2} = \mathcal{X}_m|_{y = 0} = \mathcal{X}_m|_{y = 0.2} = 30  $. At each time $t$, the image is recorded at locations $x = \frac{j}{n + 1}, y = \frac{k}{n + 1}, j, k = 1, \ldots, n$, resulting in an $n \times n $ matrix. Here, we set $ n = 21$ and $ t = 1, 2,  \ldots ,150$, which yields 150 images of size 21 $\times$ 21 for each asset. This implies that the degradation image stream of each asset can be represented by a  21 $\times$ 21 $\times$ 150 tensor. In addition, an independent and identically distributed random noise $\epsilon \sim N(0, 0.1)$ is added to each pixel. Figure \ref{fig:degradation stream images} demonstrates an example of some images with and without noise from one of the assets simulated in this study. 

To determine the TTF of an asset, we first transform the asset's $21\times 21 \times 150$ tensor to a $1\times 150$ time series by taking the average pixel intensity of each image. The time series signal indicates how the average heat of the asset involves over time. Next, we let the TTF of the asset be the time point where the amplitude of the time series signal crosses a pre-defined soft failure threshold, which is set as $23$ in this study. Since the images of different assets are generated with different thermal diffusivity coefficients, the time points where their time series signals go beyond the threshold may be different. Thus, the TTF of different asset may also be different. To mimic reality, we truncate the image stream of each asset by keeping only the images observed before its TTF. In other words, any images observed after an asset's TTF are removed from the image tensor of the asset. Such a truncation is normal in reality since an asset usually gets maintained or replaced once its degradation signal crosses the soft failure threshold. Consequently, the third dimension of the tensor of different assets might be different. In addition, to reduce the computation load, we keep one of every 10 images in the truncated image stream of each asset.

\subsection{The Benchmark and Performance Comparison}

We randomly split the generated data into a training data set consisting of $400$ assets and a test data set consisting of the remaining $100$ assets. To test the robustness of the proposed method, we consider four levels of data incompleteness: (1) $0\%$ missing, (2) $10\%$ missing, (3) $50\%$ missing, (4) $90\%$ missing. For the first scenario, (1) $0\%$ missing, we use all the generated data for model training and testing. Please notice that even though all the available images are used, the image tensor $\mathcal{X}$ is still incomplete due to failure time truncation--that is--different assets may have different TTFs and thus different number of images (see the discussion in the second paragraph of Section \ref{sec:sub:prog}). For the remaining scenarios, we randomly remove some images from each asset's image stream. For example, when $10\%$ missing, we randomly remove $10\%$ of the images (rounding to the nearest integer) from the image stream of each asset.

We compare the performance of our proposed method with an unsupervised tensor dimension reduction-based benchmark. Considering image streams are incomplete, the baseline model first applies a tensor completion method known as TMac developed by \cite{xu2013parallel} to impute the missing values of the image tensor. Next, an unsupervised tensor dimension reduction method, MPCA \citep{lu2008mpca}, is employed to reduce the dimension of the imputed image tensor to reduce dimension and extract low-dimension features, which are then used to build an LLS-based prognostic model as we discussed in Section \ref{sec:sub:prog}. MPCA is a widely used dimension reduction method for tensor data. It projects a high-dimensional tensor into a subspace but maximizes the total tensor scatter which is assumed to measure the variations in the original tensor objects. \cite{lu2008mpca} proposed a fraction-of-variation-explained (FVE) method to determine the dimension of the low-dimension tensor subspace/features, which represents the percentage of variation of the original high-dimensional tensor is preserved by the low-dimensional tensor features. Since the optimal FVE suggested by \cite{lu2008mpca} is $97\%$, we will first set FVE as $97\%$ in this study, and the corresponding baseline model is designated as ``MPCA (97\%)". In addition to the FVE method, we also use cross-validation (CV) to select an appropriate dimension for the tensor subspace. Specifically, we use the training data to conduct a 10-fold CV for various combinations of $(P_1,P_2,P_3)$, where $P_1=1,\ldots, 4$, $P_2=1,\ldots, 4$, and $P_3=1,\ldots, 4$. The baseline model is referred to as ``MPCA\_CV". We also use 10-fold CV to determine the value of the weight parameter $\alpha$ and the appropriate dimension of the tensor subspace of our proposed method. 

We use the heuristic method discussed in Section \ref{sec:sub:bua} to initialize the block updating algorithm. In this study, we use lognormal regression to build the prognostic model. The proposed method is denoted as ``Proposed\_CV". The prediction errors of our proposed method and two benchmarks are calculated by using the equation below and reported in Figures \ref{fig: complete}, \ref{fig: 10 missing}, \ref{fig: 50 missing}, and \ref{fig: 90 missing}.

\begin{equation}
\text{Prediction Error} = \frac{\vert \text{Estimated TTF} - \text{True TTF} \vert}{\text{True TTF}}.
\end{equation}

\subsection{Results and analysis}

\begin{figure*}[!htp]
\centering
\begin{minipage}[t]{0.48\textwidth}
\centering
 \includegraphics[width=1\textwidth]{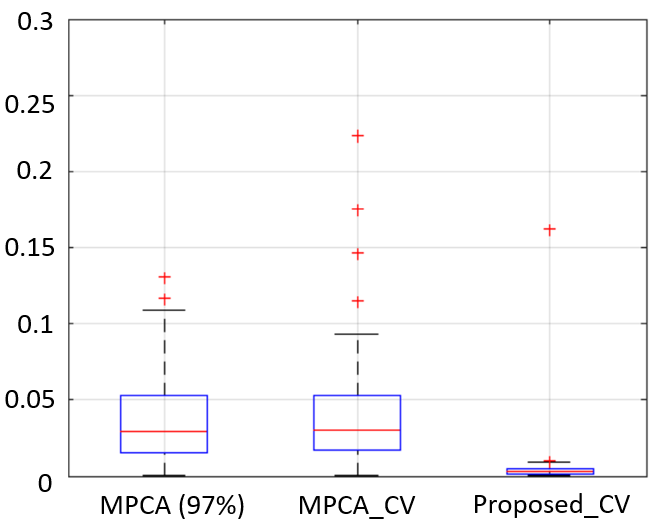}
 \caption{Prediction errors when data is complete in Numerical Study.}
 \label{fig: complete}
 \end{minipage}
 \begin{minipage}[t]{0.48\textwidth}
 \centering
 \includegraphics[width=1\textwidth]{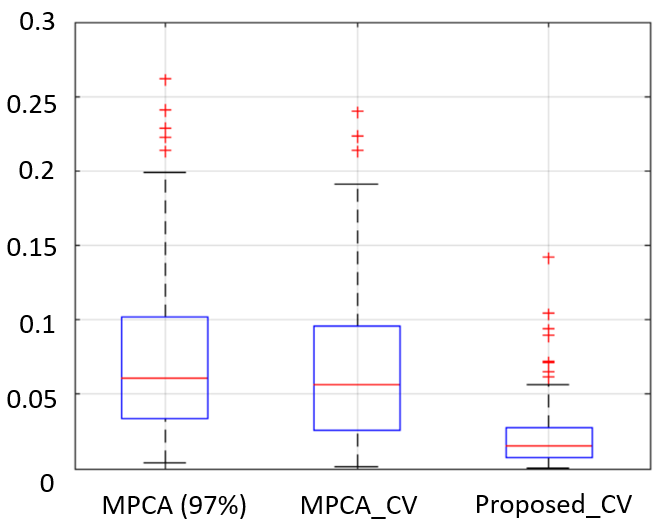}
 \caption{Prediction errors when 10\% data is missing in Numerical Study.}
 \label{fig: 10 missing}
 \end{minipage}
\end{figure*}

\begin{figure*}[!htp]
\centering

\begin{minipage}[t]{0.48\textwidth}
\centering
 \includegraphics[width=1\textwidth]{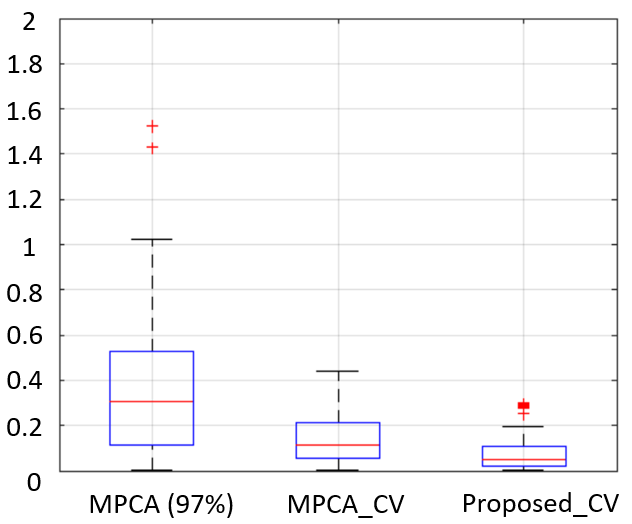}
 \caption{Prediction errors when 50\% data is missing in Numerical Study.}
 \label{fig: 50 missing}
 \end{minipage}
 \begin{minipage}[t]{0.48\textwidth}
 \centering
 \includegraphics[width=1\textwidth]{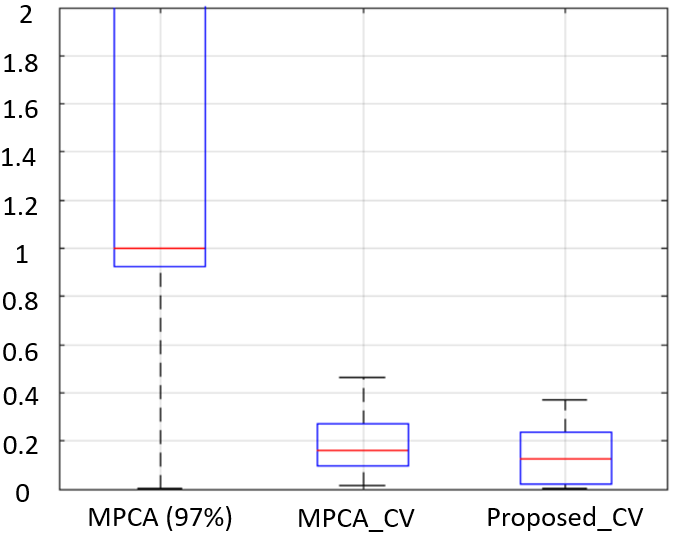}
 \caption{Prediction errors when 90\% data is missing in Numerical Study.}
 \label{fig: 90 missing}
 \end{minipage}
\end{figure*}

Figure \ref{fig: complete} reports the prediction errors of the two benchmarks and our proposed method when data is complete, which means no image is removed on purpose. Figure \ref{fig: 10 missing} shows the prediction errors when 10\% entries in the 3rd mode (time) of degradation image streams are missing, while Figures \ref{fig: 50 missing} and \ref{fig: 90 missing} demonstrate the errors when 50\% and 90\% images are missing, respectively. 

Figures \ref{fig: complete}, \ref{fig: 10 missing}, \ref{fig: 50 missing}, and \ref{fig: 90 missing} illustrate that our proposed method outperforms the benchmarks under all data missing rates. For example, when the degradation image signals are complete, the median absolute prediction errors (and the Interquartile Ranges, i.e., IQRs) of the proposed method and the two benchmarks are 0.003 (0.003), 0.027 (0.035), and 0.025 (0.033), respectively; when 10\% images are missing, the median absolute prediction errors (and IQRs) of the three methods are respectively 0.019 (0.017), 0.058 (0.067), and 0.053 (0.063); when 50\%  of images are missing, they are 0.052 (0.084), 0.302 (0.405), and 0.104 (0.168). We believe this is because our proposed method applies historical TTFs to supervise the low-dimensional tensor dimension reduction, and thus the extracted features are more effective for failure time prediction. Unlike our method, the two baseline models use MPCA, an unsupervised tensor dimension reduction method, for feature extraction. Since the extracted features are only determined by the image streams, and no TTF gets involved, they are not as effective as the features extracted by our proposed method, and thus their failure time prediction accuracy and precision are compromised.

The figures \ref{fig: complete}, \ref{fig: 10 missing}, \ref{fig: 50 missing}, and \ref{fig: 90 missing} also suggest that the performances of all the three models deteriorate, and the superiority of our proposed method over the two benchmarks decreases, with the increase of data missing rate. For example, when data is complete, the median absolute prediction errors (and IQRs) of ``Proposed\_CV" and ``MPCA\_CV" are 0.003 (0.003) and 0.025 (0.033), respectively; when the missing rate increases to 90\%, they are respectively 0.13 (0.21) and 0.16 (0.19), which are almost comparable. This is reasonable since the performance of all the models are compromised more when more data are missing. In addition, no model will perform well if a high percentage (say more than 90\%) of data are missing since it implies that very limited useful degradation information is available for modeling.

The Figures \ref{fig: complete}-\ref{fig: 90 missing} also demonstrate that ``MPCA\_CV" always outperforms ``MPCA (97\%)", and the superiority of ``MPCA\_CV" is augmented with the increase of data missing rate. For instance, when 10\% images are missing, the median absolute prediction errors (and IQR) of ``MPCA (97\%)" and ``MPCA\_CV" are 0.058 (0.067) and 0.053 (0.063), respectively; when the missing rate is 50\%, they are 0.302 (0.405) and 0.104 (0.168). One of the possible reasons is that ``MPCA (97\%)" determines the dimension of the tensor subspace by setting the ``FVE" as 97\%, which usually results in relatively high-dimensional features, although the dimension is smaller than that of the original image tensor. Relatively high-dimensional features implies an insufficient dimension reduction. In addition, it means the number of parameters in the subsequent LLS-based prognostic model is relatively large, which poses estimation challenges given that the number of samples (assets) for model training is limited.




\section{Case Study}
\label{sec:case}

\begin{figure*}[!htp]
\centering
 \includegraphics[width=0.7\textwidth]{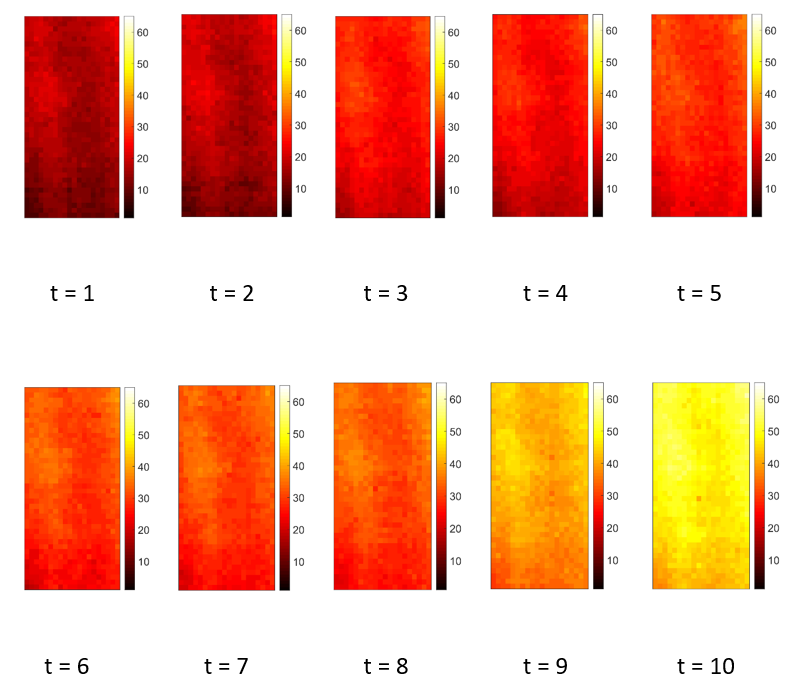}
 \caption{An illustration of one infrared degradation image stream. }
 \label{fig: Case Study_infrared}
\end{figure*}

In this section, we use degradation image streams obtained from a rotating machinery test bed to validate the effectiveness of our proposed method. The test bed is designed to perform accelerated
degradation tests on rolling entry thrust bearings. Specifically, bearings were run from brand new to failure. An FLIR T300 infrared camera was used to monitor the degradation process and collect degradation images over time. In the meanwhile, an accelerometer was mounted on the test bed to monitor the vibration of the bearing, and the failure time is defined as the time point where the amplitude of defective vibration frequencies crosses a threshold based on ISO standards for machine vibration. The data set consists of $284$ degradation image streams and their corresponding TTFs, and each image has $40\times 20$ pixels. As an illustration, a sequence of images obtained at different (ordered) time periods of one of the bearing are shown in Figure \ref{fig: Case Study_infrared}. More details about the experimental setup and the data set can be found in \cite{gebraeel2009residual} and \cite{fang2019image}.

\begin{figure*}[!htp]
\centering
\begin{minipage}[t]{0.48\textwidth}
\centering
 \includegraphics[width=1\textwidth]{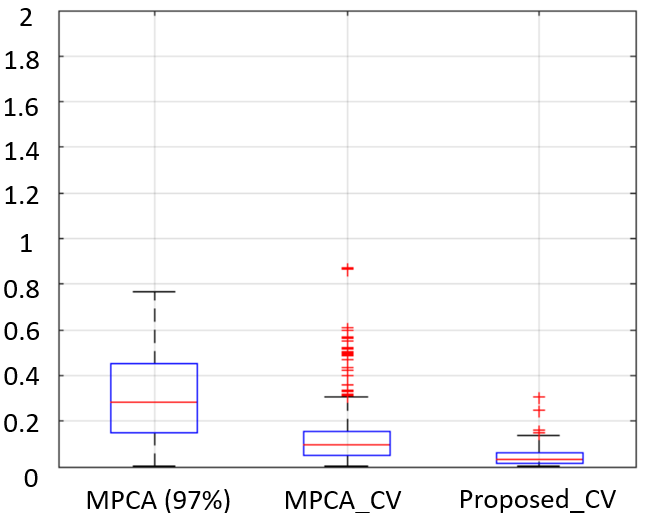}
 \caption{Prediction errors when data is complete in Case Study.}
 \label{fig: CaseStudy Complete}
 \end{minipage}
 \begin{minipage}[t]{0.48\textwidth}
 \centering
 \includegraphics[width=1\textwidth]{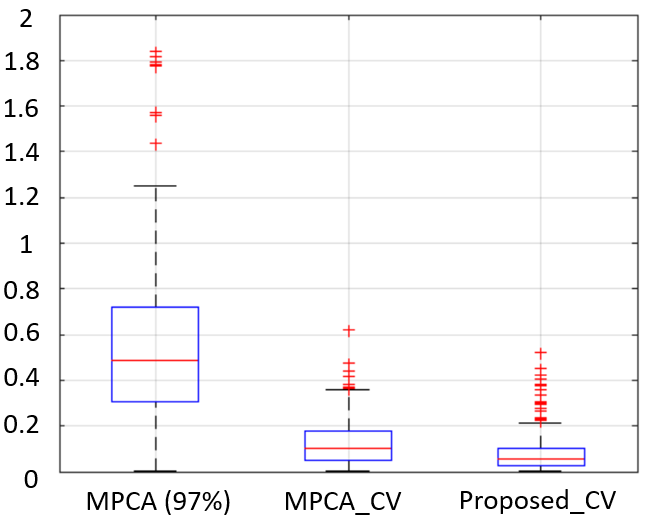}
 \caption{Prediction errors when 10\% data is missing in Case Study. }
 \label{fig: CaseStudy_Miss10}
 \end{minipage}
\end{figure*}

\begin{figure*}[!htp]
\centering
\begin{minipage}[t]{0.48\textwidth}
\centering
 \includegraphics[width=1\textwidth]{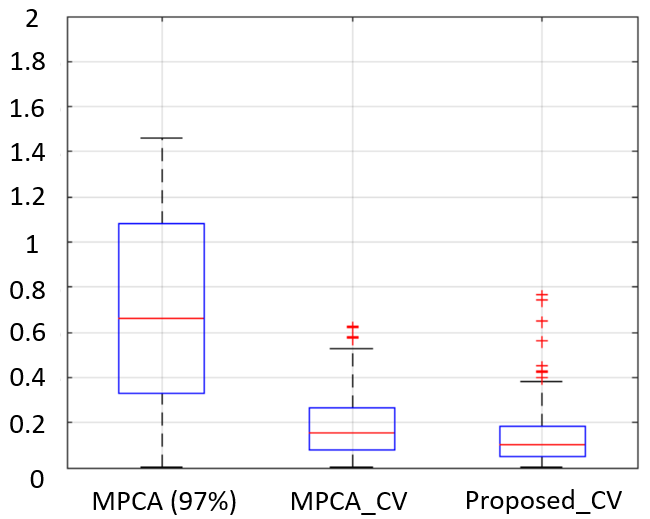}
 \caption{Prediction errors when 50\% data is missing in Case Study. }
 \label{fig: CaseStudy_Miss50}
 \end{minipage}
 \begin{minipage}[t]{0.48\textwidth}
 \centering
 \includegraphics[width=1\textwidth]{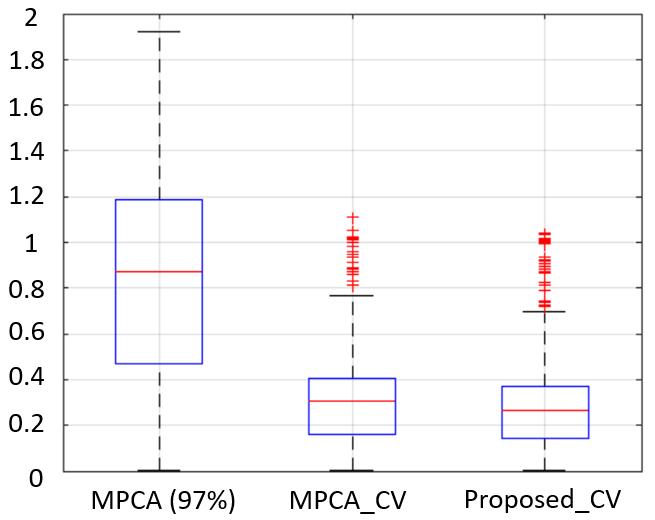}
 \caption{Prediction errors when 90\% data is missing in Case Study. }
 \label{fig: CaseStudy_Miss90}
 \end{minipage}
\end{figure*}

We use 5-fold cross validation to evaluate the performance of our proposed model and the two benchmarks discussed in Section \ref{sec:numer}. Similar to the simulation study, we conduct 10-fold cross validation to determine the optimal weight parameter in criterion \eqref{eq:supervised} and the most appropriate dimension of the tensor subspace. In addition, we also consider four levels of data incompleteness: (1) $0\%$ missing (i.e., complete), (2) $10\%$ missing, (3) $50\%$ missing, and (4) $90\%$ missing. Figure \ref{fig: CaseStudy Complete} illustrates the absolute prediction errors when degradation image streams are complete. Figure \ref{fig: CaseStudy_Miss10} shows prediction errors when 10\% of the images of each bearing are missing. Figures \ref{fig: CaseStudy_Miss50} and \ref{fig: CaseStudy_Miss90}  demonstrate the absolute prediction errors when the missing rate are 50\% and 90\%, respectively.

Similar to the discovery in the numerical study in Section \ref{sec:numer}, Figures \ref{fig: CaseStudy Complete}, \ref{fig: CaseStudy_Miss10}, \ref{fig: CaseStudy_Miss50}, and \ref{fig: CaseStudy_Miss90} indicate that our proposed method constantly works better than the two benchmarks under all the 4 data missing rates. For example, the median absolute prediction errors (and IQRs) of our proposed method and the two benchmarks are 0.03 (0.04), 0.3 (0.24), and 0.1 (0.16), respectively, when the degradation image streams are complete. When 50\% of the images are missing, the median absolute prediction errors (and IQR) of are respectively 0.09 (0.17), 0.62 (0.65), and 0.15 (0.19). We believe this is because our proposed model is a supervised dimension reduction-based method, which uses TTF information to supervise the defection of the low-dimensional tensor subspace, while the benchmarks are unsupervised dimension reduction-based methods without TTF information involved. Since our method considers TTF information when detecting the tensor subspace, the extracted features are more effective for failure time prediction.

Figure \ref{fig: CaseStudy Complete}, \ref{fig: CaseStudy_Miss10}, \ref{fig: CaseStudy_Miss50}, \ref{fig: CaseStudy_Miss90} also show that the prediction errors of all the 3 methods increase with the increase of data missing rates. For example, when the missing rates are 0\%, 10\%, 50\%, and 90\%, the median absolute prediction errors (and IQRs) of ``MPCA (97\%)" are 0.3 (0.24), 0.49 (0.42), 0.62 (0.65), and 0.83 (0.78), respectively, while they are respectively 0.09 (0.11), 0.1 (0.16), 0.15 (0.19), and 0.31 (0.22) for ``MPCA\_CV", and 0.03 (0.04), 0.05 (0.08), 0.09 (0.17), and 0.29 (0.21) for our proposed method. This is reasonable since a higher data missing rate means less useful degradation information and thus a worse model performance. In addition, we observe that the superiority of our proposed method over the two benchmarks decrease with the increase of data missing rates. For example, the prediction accuracy of our method and ``MPCA (97\%)" are comparable when the data missing rate is 90\%. We again believe this is because not much useful information is available when data is highly incomplete and none of the two models perform well with such limited data.

We also observe that ``MPCA\_CV" always outperforms ``MPCA (97\%)", and the superiority of ``MPCA\_CV" is augmented with the increase of data missing rate. For example, when 10\% images are missing, the median absolute prediction errors (and IQR) of ``MPCA (97\%)" and MPCA\_CV are 0.49 (0.42) and 0.1 (0.16), respectively; when the missing rate is 50\%, the they are 0.62 (0.65) and 0.15 (0.19). We again believe this is because ``MPCA (97\%)" determines the dimension of the tensor subspace by setting the ``FVE" as 97\%, which  results in relatively high-dimensional features due to the insufficient dimension reduction. Also, it results in a parameter estimation challenge since the number of parameters to be estimated in the prognostic model is relatively large comparing to the limited number of historical samples for model training. This suggests that cross validation is a better method to determine the dimension of the tensor subspace, especially when we do not have enough number of samples for model training.

\section{Conclusions}
\label{sec:concl}
This paper proposed a supervised tensor dimension reduction-based prognostic model for applications with incomplete degradation imaging data. This is achieved by first developing a new supervised tensor dimension reduction method that reduces the dimension of incomplete high-dimensional degradation image streams and provides low-dimensional tensor features, which are then used to build a prognostic model 
based on (log)-location-scale regression.

The supervised tensor dimension reduction method uses historical TTFs to supervise the detection of a low-dimensional tensor subspace to reduce the dimension of incomplete high-dimensional image streams. Mathematically, it is formulated as an optimization criterion that combines a feature extraction term and a regression term. The feature extraction term focuses on identifying a tensor space to extract low-dimensional tensor features from high-dimensional image streams. The regression term regresses failure times against the features extracted by the first term using LLS regression. By jointly optimizing the two terms, it is expected to detect an appropriate tensor subspace such that the extracted features are effective for TTF prediction. To estimate the parameters of the supervised dimension reduction method, we developed a Block Updating Algorithm for applications where TTFs follow distributions in the (log)-location-scale family. The algorithm works by splitting the parameters into several blocks and cyclically optimizing one block of parameters while keeping other blocks fixed until convergence. In addition, we showed that if TTFs follow normal or lognormal distributions, there is a closed-form solution when optimizing each block of the parameters, no matter the imaging data is complete or incomplete. 

Simulated data as well as a data set from rotating machinery were used to validate the effectiveness of our proposed method. The results showed that our proposed prognostic method consistently outperforms the unsupervised tensor reduction-based benchmarks under various data missing rates. This validated the benefits and importance of using failure time information to supervise the dimension reduction of high-dimensional degradation image streams when building prognostic models. 

The proposed prognostic model assumes that the TTFs of assets in the training dataset are known. In many real-world applications, the historical failure times might be right censored. This is because a component might be replaced before failure, so the exact TTF is unknown, and we only know that it is larger than the replacement time. How to incorporate censored TTFs into the proposed method can be an interesting future research topic.

\newpage

~~~~~~~~~~~~~~~~~~~~~~~~~~~~~~~~~~~~~~~~~~~~~\textbf{Appendix}
\setcounter{section}{0}
\section{Proof of Proposition 1}

The original optimization problem is 
\begin{equation*}
\arg\min_{\bm{U}_{1}}~~\alpha||\mathcal{P}_{\Omega}(\mathcal{X} - \mathcal{S} \times_{1} \bm{U}_{1}^\top \times_{2} \bm{U}_{2}^\top \times_{3} \bm{U}_{3}^\top )||_{F}^2 + (1 - \alpha)||\bm{y} - \bm{1}_{M} \cdot \tilde{\beta}_{0} - \bm{S}_{(4)} \cdot \tilde{\bm{\beta}}_{1}||_{F}^{2},
\end{equation*}

\noindent which is equivalent to the following problem when data is complete:
\begin{equation*}
\arg\min_{\bm{U}_{1}}~~\alpha||\mathcal{X} - \mathcal{S} \times_{1} \bm{U}_{1}^\top \times_{2} \bm{U}_{2}^\top \times_{3} \bm{U}_{3}^\top ||_{F}^2 + (1 - \alpha)||\bm{y} - \bm{1}_{M} \cdot \tilde{\beta}_{0} - \bm{S}_{(4)} \cdot \tilde{\bm{\beta}}_{1}||_{F}^{2},
\end{equation*}

\noindent which is convex. Thus, it can be solved by setting the derivatives to be zeros--that is $\frac{d\Psi}{d\bm{U}_{1}} = \bm{0}$, where $\Psi=\alpha||\mathcal{X} - \mathcal{S} \times_{1} \bm{U}_{1}^\top \times_{2} \bm{U}_{2}^\top \times_{3} \bm{U}_{3}^\top ||_{F}^2 + (1 - \alpha)||\bm{y} - \bm{1}_{M} \cdot \tilde{\beta}_{0} - \bm{S}_{(4)} \cdot \tilde{\bm{\beta}}_{1}||_{F}^{2}$. This implies $\frac{d}{d\bm{U}_{1}}(||\mathcal{X} - \mathcal{S} \times_{1} \bm{U}_{1}^\top \times_{2} \bm{U}_{2}^\top \times_{3} \bm{U}_{3}^\top ||_{F}^2) = \bm{0}$. According to the communication law of tensor mode multiplication, we have $\frac{d}{d\bm{U}_{1}}(||\mathcal{X} - (\mathcal{S} \times_{2} \bm{U}_{2}^\top \times_{3} \bm{U}_{3}^\top) \times_{1} \bm{U}_{1}^\top ||_{F}^2) = \bm{0}$. Thus, $\frac{d}{d\bm{U}_{1}}(||\mathcal{X} - \mathcal{S}_{{U}_{1}} \times_{1} \bm{U}_{1}^\top ||_{F}^2) = \bm{0}$, where $\mathcal{S}_{{U}_{1}}=\mathcal{S} \times_{2} \bm{U}_{2}^\top \times_{3} \bm{U}_{3}^\top$. Furthermore, we have $\frac{d}{d\bm{U}_{1}}(||\bm{X}_{(1)} - \bm{U}_{1}^\top  \cdot \bm{S}_{{U}_{1}(1)} ||_{F}^2) = \bm{0}$ due to the fact that $||\mathcal{S}||_{F}^{2} = ||\bm{S}_{(n)}||_{F}^{2}$ and the property of tensor mode multiplication $\mathcal{S} \times_{n} \bm{U} = \bm{U} \cdot \bm{S}_{(n)}$. By taking the derivative of the Frobenius norm, we have $2(\bm{X}_{(1)} - \bm{U}_{1}^\top  \cdot \bm{S}_{{U}_{1}(1)}) \cdot (-\bm{S}_{{U}_{1}(1)}^\top) = \bm{0}$. Thus, $\bm{U}_{1}^\top \cdot \bm{S}_{{U}_{1}(1)} \cdot \bm{S}_{{U}_{1}(1)}^\top = \bm{X}_{(1)} \cdot \bm{S}_{{U}_{1}(1)}^\top$, which gives that $\bm{U}_{1}^\top = \bm{X}_{(1)} \cdot \bm{S}_{{U}_{1}(1)}^\top \cdot ( \bm{S}_{{U}_{1}(1)} \cdot \bm{S}_{{U}_{1}(1)}^\top)^{-1}$. Finally, we have $\bm{U}_{1} = (\bm{X}_{(1)} \cdot \bm{S}_{{U}_{1}(1)}^\top \cdot ( \bm{S}_{{U}_{1}(1)} \cdot \bm{S}_{{U}_{1}(1)}^\top)^{-1})^\top$.

\section{Proof of Proposition 2}

The original optimization problem is
\begin{equation*}
\arg \min_{\bm{U}_{2}}~~\alpha||\mathcal{P}_{\Omega}(\mathcal{X} - \mathcal{S} \times_{1} \bm{U}_{1}^\top \times_{2} \bm{U}_{2}^\top \times_{3} \bm{U}_{3}^\top )||_{F}^2 + (1 - \alpha)||\bm{y} - \bm{1}_{M} \cdot \tilde{\beta}_{0} - \bm{S}_{(4)} \cdot \tilde{\bm{\beta}}_{1}||_{F}^{2} \mbox{},
\end{equation*}

\noindent which is equivalent to the following problem when data is complete:
\begin{equation*}
\arg \min_{\bm{U}_{2}}~~\alpha||\mathcal{X} - \mathcal{S} \times_{1} \bm{U}_{1}^\top \times_{2} \bm{U}_{2}^\top \times_{3} \bm{U}_{3}^\top ||_{F}^2 + (1 - \alpha)||\bm{y} - \bm{1}_{M} \cdot \tilde{\beta}_{0} - \bm{S}_{(4)} \cdot \tilde{\bm{\beta}}_{1}||_{F}^{2} \mbox{}
\end{equation*}

\noindent which is convex. Therefore, it can be solved by setting the derivatives to be zeros--that is 
$\frac{d\Psi}{d\bm{U}_{2}} = \bm{0}$, where $\Psi = \alpha||\mathcal{X} - \mathcal{S} \times_{1} \bm{U}_{1}^\top \times_{2} \bm{U}_{2}^\top \times_{3} \bm{U}_{3}^\top ||_{F}^2 + (1 - \alpha)||\bm{y} - \bm{1}_{M} \cdot \tilde{\beta}_{0} - \bm{S}_{(4)} \cdot \tilde{\bm{\beta}}_{1}||_{F}^{2}$. This implies $\frac{d}{d\bm{U}_{2}}(||\mathcal{X} - \mathcal{S} \times_{1} \bm{U}_{1}^\top \times_{2} \bm{U}_{2}^\top \times_{3} \bm{U}_{3}^\top ||_{F}^2) = \bm{0}$. According to the communication law of tensor mode multiplication, we have $\frac{d}{d\bm{U}_{2}}(||\mathcal{X} - (\mathcal{S} \times_{1} \bm{U}_{1}^\top \times_{3} \bm{U}_{3}^\top) \times_{2} \bm{U}_{2}^\top ||_{F}^2) = \bm{0}$. Thus, $\frac{d}{d\bm{U}_{2}}(||\mathcal{X} - \mathcal{S}_{{U}_{2}} \times_{2} \bm{U}_{2}^\top ||_{F}^2) = \bm{0}$, where $\mathcal{S}_{{U}_{2}}=\mathcal{S} \times_{1} \bm{U}_{1}^\top \times_{3} \bm{U}_{3}^\top$. Furthermore, we have $\frac{d}{d\bm{U}_{2}}(||\bm{X}_{(2)} - \bm{U}_{2}^\top  \cdot \bm{S}_{{U}_{2}(2)} ||_{F}^2) = \bm{0}$ due to the fact that $||\mathcal{S}||_{F}^{2} = ||\bm{S}_{(n)}||_{F}^{2}$ and the property of tensor mode multiplicaion $\mathcal{S} \times_{n} \bm{U} = \bm{U} \cdot \bm{S}_{(n)}$. By taking the derivative of the Frobenius norm, we have $2(\bm{X}_{(2)} - \bm{U}_{2}^\top  \cdot \bm{S}_{{U}_{2}(2)}) \cdot (-\bm{S}_{{U}_{2}(2)}^\top) = \bm{0}$. Thus, $\bm{U}_{2}^\top \cdot \bm{S}_{{U}_{2}(2)} \cdot \bm{S}_{{U}_{2}(2)}^\top = \bm{X}_{(2)} \cdot \bm{S}_{{U}_{2}(2)}^\top$ which gives that $\bm{U}_{2}^\top = \bm{X}_{(2)} \cdot \bm{S}_{{U}_{2}(2)}^\top \cdot ( \bm{S}_{{U}_{2}(2)} \cdot \bm{S}_{{U}_{2}(2)}^\top)^{-1}$. Finally we have $\bm{U}_{2} = (\bm{X}_{(2)} \cdot \bm{S}_{{U}_{2}(2)}^\top \cdot ( \bm{S}_{{U}_{2}(2)} \cdot \bm{S}_{{U}_{2}(2)}^\top)^{-1})^\top$.

\section{Proof of Proposition 3}
The original optimization problem is
\begin{equation*}
\arg \min_{\bm{U}_{3}}~~\alpha||\mathcal{P}_{\Omega}(\mathcal{X} - \mathcal{S} \times_{1} \bm{U}_{1}^\top \times_{2} \bm{U}_{2}^\top \times_{3} \bm{U}_{3}^\top )||_{F}^2 + (1 - \alpha)||\bm{y} - \bm{1}_{M} \cdot \tilde{\beta}_{0} - \bm{S}_{(4)} \cdot \tilde{\bm{\beta}}_{1}||_{F}^{2},
\end{equation*}

\noindent which is equivalent to the following problem when data is complete:
\begin{equation*}
\arg \min_{\bm{U}_{3}}~~\alpha||\mathcal{X} - \mathcal{S} \times_{1} \bm{U}_{1}^\top \times_{2} \bm{U}_{2}^\top \times_{3} \bm{U}_{3}^\top ||_{F}^2 + (1 - \alpha)||\bm{y} - \bm{1}_{M} \cdot \tilde{\beta}_{0} - \bm{S}_{(4)} \cdot \tilde{\bm{\beta}}_{1}||_{F}^{2}
\end{equation*}

\noindent which is convex. Therefore, it can be solved by setting the derivatives to be zeros--that is $ \frac{d\Psi}{d\bm{U}_{3}} = \bm{0}$, where  $\Psi = \alpha||\mathcal{X} - \mathcal{S} \times_{1} \bm{U}_{1}^\top \times_{2} \bm{U}_{2}^\top \times_{3} \bm{U}_{3}^\top ||_{F}^2 + (1 - \alpha)||\bm{y} - \bm{1}_{M} \cdot \tilde{\beta}_{0} - \bm{S}_{(4)} \cdot \tilde{\bm{\beta}}_{1}||_{F}^{2}$. This implies $\frac{d}{d\bm{U}_{3}}(||\mathcal{X} - \mathcal{S} \times_{1} \bm{U}_{1}^\top \times_{2} \bm{U}_{2}^\top \times_{3} \bm{U}_{3}^\top ||_{F}^2) = \bm{0}$. According to the communication law of tensor mode multiplication, we have $\frac{d}{d\bm{U}_{3}}(||\mathcal{X} - (\mathcal{S} \times_{1} \bm{U}_{1}^\top \times_{2} \bm{U}_{2}^\top) \times_{3} \bm{U}_{3}^\top ||_{F}^2) = \bm{0}$. Thus, $ \frac{d}{d\bm{U}_{3}}(||\mathcal{X} - \mathcal{S}_{{U}_{3}} \times_{3} \bm{U}_{3}^\top ||_{F}^2) = \bm{0}$, where $\mathcal{S}_{{U}_{3}} = \mathcal{S} \times_{1} \bm{U}_{1}^\top \times_{2} \bm{U}_{2}^\top$. Furthermore, we have $\frac{d}{d\bm{U}_{3}}(||\bm{X}_{(3)} - \bm{U}_{3}^\top  \cdot \bm{S}_{{U}_{3}(3)} ||_{F}^2) = \bm{0}$ due to the fact that $||\mathcal{S}||_{F}^{2} = ||\bm{S}_{(n)}||_{F}^{2}$ and the property of tensor mode multiplicaion $\mathcal{S} \times_{n} \bm{U} = \bm{U} \cdot \bm{S}_{(n)}$. By taking the derivative of the Frobenius norm, we have $ 2(\bm{X}_{(3)} - \bm{U}_{3}^\top  \cdot \bm{S}_{{U}_{3}(3)}) \cdot (-\bm{S}_{{U}_{3}(3)}^\top) = \bm{0}$. Thus, $\bm{U}_{3}^\top \cdot \bm{S}_{{U}_{3}(3)} \cdot \bm{S}_{{U}_{3}(3)}^\top = \bm{X}_{(3)} \cdot \bm{S}_{{U}_{3}(3)}^\top$, which gives that  $\bm{U}_{3}^\top = \bm{X}_{(3)} \cdot \bm{S}_{{U}_{3}(3)}^\top \cdot ( \bm{S}_{{U}_{3}(3)} \cdot \bm{S}_{{U}_{3}(3)}^\top)^{-1}$. Finally, we have $\bm{U}_{3} = (\bm{X}_{(3)} \cdot \bm{S}_{{U}_{3}(3)}^\top \cdot ( \bm{S}_{{U}_{3}(3)} \cdot \bm{S}_{{U}_{3}(3)}^\top)^{-1})^\top$.

\section{Proof of Proposition 4}
The original optimization problem is 
\begin{equation*}
\arg \min_{S}~~\alpha||\mathcal{P}_{\Omega}(\mathcal{X} - \mathcal{S} \times_{1} \bm{U}_{1}^\top \times_{2} \bm{U}_{2}^\top \times_{3} \bm{U}_{3}^\top )||_{F}^2 + (1 - \alpha)||\bm{y} - \bm{1}_{M} \cdot \beta_{0} - \bm{S}_{(4)} \cdot \bm{\beta}_{1}||_{F}^{2},
\end{equation*}

\noindent which is equivalent to the following problem when data is complete:
\begin{equation*}
\arg \min_{\hat{S}}~~\alpha||\mathcal{X} - \mathcal{S} \times_{1} \bm{U}_{1}^\top \times_{2} \bm{U}_{2}^\top \times_{3} \bm{U}_{3}^\top ||_{F}^2 + (1 - \alpha)||\bm{y} - \bm{1}_{M} \cdot \beta_{0} - \bm{S}_{(4)} \cdot \bm{\beta}_{1}||_{F}^{2},
\end{equation*}

\noindent which is convex. Thus, it can be solved by setting the derivatives to be zeros--that is 
$\frac{d\Psi}{d{\bm{S}}} = \bm{0}$, where $\Psi = \alpha||\mathcal{X} - \mathcal{S} \times_{1} \bm{U}_{1}^\top \times_{2} \bm{U}_{2}^\top \times_{3} \bm{U}_{3}^\top ||_{F}^2 + (1 - \alpha)||\bm{y} - \bm{1}_{M} \cdot \beta_{0} - \bm{S}_{(4)} \cdot \bm{\beta}_{1}||_{F}^{2}$. According to the connection between Kronecker product and tensor mode multiplication \citep{kolda2006multilinear}, we have $ \frac{d}{d{\bm{S}}}(\alpha||\bm{X}_{(4)} - \bm{S}_{(4)} \cdot (\bm{U}_{3} \otimes \bm{U}_{2} \otimes \bm{U}_{1}) ||_{F}^2 + (1 - \alpha)||\bm{y} - \bm{1}_{M}\cdot\beta_{0} - \bm{S}_{(4)} \cdot \bm{\beta}_{1}||_{F}^{2}) = \bm{0}$. By taking the derivative of the Frobenius norm, we have $2\alpha \cdot [\bm{X}_{(4)} - \bm{S}_{(4)} \cdot (\bm{U}_{3} \otimes \bm{U}_{2} \otimes \bm{U}_{1})]\cdot [- (\bm{U}_{3} \otimes \bm{U}_{2} \otimes \bm{U}_{1})^\top] + 2(1 - \alpha) \cdot [\bm{y} - \bm{1}_{M} \cdot \beta_{0} - \bm{S}_{(4)} \cdot \bm{\beta}_{1}] \cdot (-\bm{\beta}_{1}^\top) = \bm{0}$. Thus, $ -2\alpha \cdot \bm{X}_{(4)} \cdot (\bm{U}_{3} \otimes \bm{U}_{2} \otimes \bm{U}_{1})^\top + 2\alpha \cdot \bm{S}_{(4)} \cdot (\bm{U}_{3} \otimes \bm{U}_{2} \otimes \bm{U}_{1}) \cdot (\bm{U}_{3} \otimes \bm{U}_{2} \otimes \bm{U}_{1})^\top + 2(1-\alpha) (\bm{y} - \bm{1}_{M} \cdot \beta_{0})\cdot (-\bm{\beta}_{1}^\top) + 2(1 - \alpha) \cdot (\bm{S}_{(4)} \cdot \bm{\beta}_{1} \cdot \bm{\beta}_{1}^\top) = \bm{0}$. Finally, we have $\bm{S}_{(4)} = \big[ \alpha \cdot \bm{X}_{(4)} \cdot (\bm{U}_{3} \otimes \bm{U}_{2} \otimes \bm{U}_{1})^\top + (1 - \alpha) \cdot (\bm{y} - \bm{1}_{M} \cdot \beta_{0}) \cdot \bm{\beta}_{1}^\top \big] \cdot \big[ \alpha \cdot (\bm{U}_{3} \otimes \bm{U}_{2} \otimes \bm{U}_{1}) \cdot (\bm{U}_{3} \otimes \bm{U}_{2} \otimes \bm{U}_{1})^\top + (1 - \alpha) \cdot \bm{\beta}_{1} \cdot \bm{\beta}_{1}^\top \big]^{-1}$.

\section {Proof of Lemma 1}

Let $\bm{a}_{m} \in \mathbb{R}^{1 \times N}$ denotes the $m$th row of matrix $\bm{A} \in \mathbb{R}^{M \times N}$ and $\bm{b}_{m} \in \mathbb{R}^{1 \times P}$ denotes the $m$th row of matrix $\bm{B} \in \mathbb{R}^{M \times P}$, $m = 1, \ldots, M$, then we have 
\begin{equation*}
\bm{A} - \bm{B}\bm{C} = [(\bm{a}_{1} - \bm{b}_{1}\bm{C})^\top, \ldots, (\bm{a}_{M} - \bm{b}_{M}\bm{C})^\top ]^\top.
\end{equation*}

\noindent Based on the definition of Frobenius norm, the original objective function in  Lemma \ref{lemma1} can be transformed as follows:
\begin{equation*}
\|\bm{A} - \bm{B}\bm{C} \|_{F}^{2} = \|[(\bm{a}_{1} - \bm{b}_{1}\bm{C})^\top, \ldots, (\bm{a}_{M} - \bm{b}_{M}\bm{C})^\top ]^\top \|_{F}^{2} = \sum_{m = 1}^{M} \| \bm{a}_{m} - \bm{b}_{m}\bm{C}\|_{F}^{2}.
\end{equation*}

\noindent Therefore, we have the following:
\begin{equation*}
\arg \min_{\bm{B}} \| \bm{A} - \bm{B}\bm{C}\|_{F}^{2} = \arg \min_{\{\bm{b}_{m}\}_{m=1}^M}\sum_{m = 1}^{M} \| \bm{a}_{m} - \bm{b}_{m}\bm{C}\|_{F}^{2}.
\end{equation*}

\noindent where ${\bm{B} = [\bm{b}_{1}^\top, \ldots, \bm{b}_{M}^\top]^\top}$. Therefore, to solve the original objective function, we can simply solve the following $M$ sub problems:
\begin{equation*}
\arg \min_{\bm{b}_{m}}\| \bm{a}_{m} - \bm{b}_{m}\bm{C}\|_{F}^{2}, \quad m = 1, \ldots, M.
\end{equation*}


\section{Proof of Proposition 5}

The original optimization problem is 
\begin{equation*}
\arg \min_{\bm{U}_{1}}~~\alpha||\mathcal{P}_{\Omega}(\mathcal{X} - \mathcal{S} \times_{1} \bm{U}_{1}^\top \times_{2} \bm{U}_{2}^\top \times_{3} \bm{U}_{3}^\top )||_{F}^2 + (1 - \alpha)||\bm{y} - \bm{1}_{M} \cdot \tilde{\beta}_{0} - \bm{S}_{(4)} \cdot \tilde{\bm{\beta}}_{1}||_{F}^{2},
\end{equation*}

\noindent which is equivalent to the following problem when data is missing:
\begin{equation*}
\arg \min_{\bm{U}_{1}}~~\alpha||\mathcal{X} - (\mathcal{S} \times_{1} \bm{U}_{1}^\top \times_{2} \bm{U}_{2}^\top \times_{3} \bm{U}_{3}^\top )\odot \text{logic}(\mathcal{X})||_{F}^2 + (1 - \alpha)||\bm{y} - \bm{1}_{M} \cdot \tilde{\beta}_{0} - \bm{S}_{(4)} \cdot \tilde{\bm{\beta}}_{1}||_{F}^{2},
\end{equation*}

\noindent where $\odot$ is the inner product, \text{logic}($\mathcal{X}$) denotes the logical value of $\mathcal{X}$--that is--if an entry is observed, its logical value is 1; otherwise, it is 0. Since the problem is convex, it can be solved by setting the derivatives to be zeros, i.e., $ \frac{d\Psi}{d\bm{U}_{1}} = \bm{0}$,  where $\Psi = \alpha||\mathcal{X} - (\mathcal{S} \times_{1} \bm{U}_{1}^\top \times_{2} \bm{U}_{2}^\top \times_{3} \bm{U}_{3}^\top )\odot \text{logic}(\mathcal{X})||_{F}^2 + (1 - \alpha)||\bm{y} - \bm{1}_{M} \cdot \tilde{\beta}_{0} - \bm{S}_{(4)} \cdot \tilde{\bm{\beta}}_{1}||_{F}^{2}$. This implies 
$ \frac{d}{d\bm{U}_{1}}(||\mathcal{X} - (\mathcal{S} \times_{1} \bm{U}_{1}^\top \times_{2} \bm{U}_{2}^\top \times_{3} \bm{U}_{3}^\top) \odot \text{logic}(\mathcal{X}) ||_{F}^2) = \bm{0}$. According to the communication law of tensor mode multiplication, we have $\frac{d}{d\bm{U}_{1}}(||\mathcal{X} - [(\mathcal{S} \times_{1} \bm{U}_{1}^\top \times_{2} \bm{U}_{2}^\top) \times_{3} \bm{U}_{3}^\top] \odot \text{logic}(\mathcal{X}) ||_{F}^2) = \bm{0}$. Thus, $\frac{d}{d\bm{U}_{1}} (||\mathcal{X} - (\mathcal{S}_{{U}_{1}} \times_{1} \bm{U}_{1}^\top )\odot \text{logic}(\mathcal{X})||_{F}^2)= \bm{0}$, where $\mathcal{S}_{{U}_{1}} = \mathcal{S} \times_{2} \bm{U}_{2}^\top \times_{3} \bm{U}_{3}^\top$. Furthermore, we have $\frac{d}{d\bm{U}_{1}}(||\bm{X}_{(1)} - (\bm{U}_{1}^\top  \cdot \bm{S}_{{U}_{1}(1)})\odot \text{logic}(\bm{X}_{(1)}) ||_{F}^2) = \bm{0}$ since $||\mathcal{S}||_{F}^{2} = ||\bm{S}_{(n)}||_{F}^{2}$ and $\mathcal{S} \times_{n} \bm{U} = \bm{U} \cdot \bm{S}_{(n)}$.

\begin{figure*}[!htp]
\centering
 \includegraphics[width=0.8\textwidth]{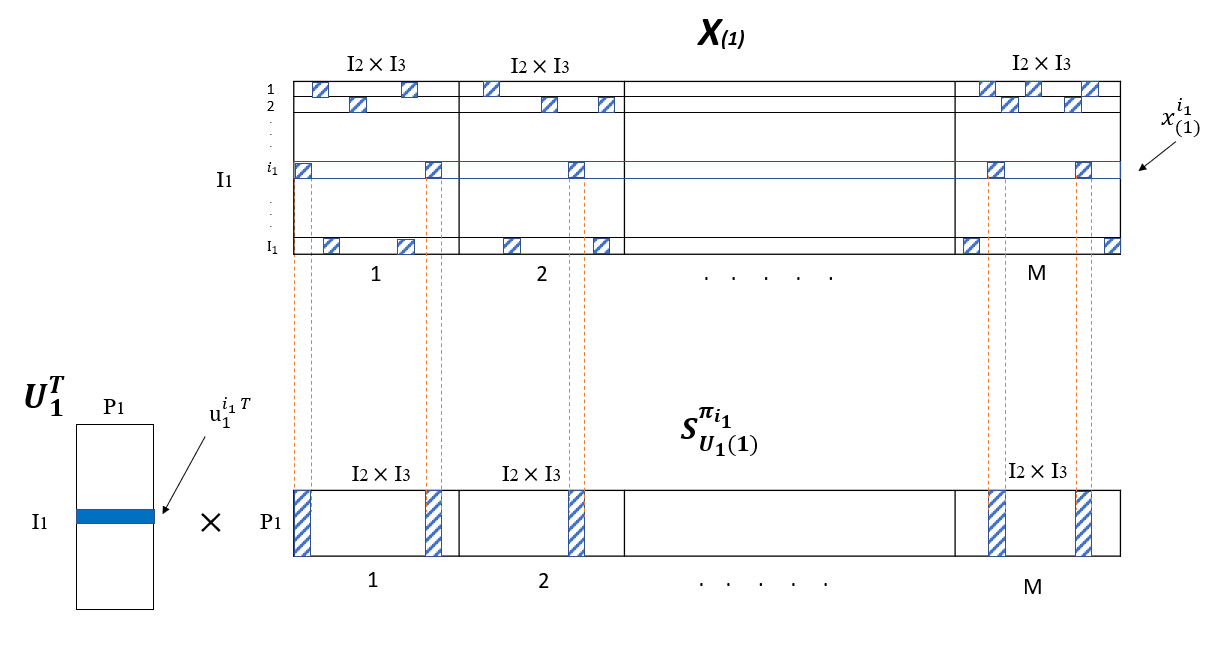}
 \caption{An illustration of the data missing pattern in Proposition 5 (stripes representing available entries).}
 \label{fig: U1 Entry Missing}
\end{figure*}

Figure \ref{fig: U1 Entry Missing} shows the pattern of mode-1 matricization of the 4D tensor $\mathcal{X}$ when it has missing entries whose indices can be denoted by a set $\Omega \subseteq \{ (i_1, i_2, i_3, m),1\leq i_1\leq I_1, 1\leq i_2\leq I_2,1\leq i_3\leq I_3,1\leq m\leq M\}$. Based on Lemma \ref{lemma1}, we can sequentially optimize each column of $\bm{U}_{1}$.

Specifically, we denote the $i_1$th row in $\bm{U}_{1}^\top$ as $\bm{u}_{1}^{i_1 \top}$ (blue solid row of $\bm{U}_{1}^\top$ in Figure \ref{fig: U1 Entry Missing}). The available entries in the $i_1$th row of $\bm{X}_{(1)}$ are denoted as $\bm{x}_{(1)}^{i_1,\pi_{i_1}}$ (blue striped squares of $\bm{x}_{(1)}^{i_1}$ in Figure \ref{fig: U1 Entry Missing}). In $\bm{S}_{U_1(1)}^{\pi_{i_1}}$, we choose the columns whose indices are the same as those of the available entries of $\bm{x}_{(1)}^{i_1}$ (blue striped columns of $\bm{S}_{U_1(1)}^{\pi_{i_1}}$ in Figure \ref{fig: U1 Entry Missing}). As a result, we have $\frac{d}{d\bm{u}_{1}^{i_1}}(\Sigma_{{i_1} = 1}^{I_1}||\bm{x}_{(1)}^{i_1,\pi_{i_1}} - ({\bm{u}_{1}^{i_1 \top}}  \cdot \bm{S}_{{U}_{1}(1)}^{\pi_{i_1}})||_{F}^2) = \bm{0}$. Because we only take the derivative of $\bm{u}_{1}^{i_1}$, we have $\frac{d}{d\bm{u}_{1}^{i_1}}(||\bm{x}_{(1)}^{i_1,\pi_{i_1}} - ({\bm{u}_{1}^{i_1 \top}}  \cdot \bm{S}_{{U}_{1}(1)}^{\pi_{i_1}})||_{F}^2) = \bm{0}$. By taking the derivative of the Frobenius norm, we have
$2(\bm{x}_{(1)}^{i_1,\pi_{i_1}} - \bm{u}_{1}^{i_1 \top}  \cdot \bm{S}_{{U}_{1}(1)}^{\pi_{i_1}}) \cdot (-{\bm{S}_{{U}_{1}(1)}^{\pi_{i_1}}}^\top) = \bm{0}$. Thus, $\bm{u}_{1}^{i_1 \top} \cdot \bm{S}_{{U}_{1}(1)}^{\pi_{i_1}} \cdot {\bm{S}_{{U}_{1}(1)}^{\pi_{i_1}}}^\top = \bm{x}_{(1)}^{i_1,\pi_{i_1}} \cdot {\bm{S}_{{U}_{1}(1)}^{\pi_{i_1}}}^\top$, which gives that $\bm{u}_{1}^{i_1} = (\bm{x}_{(1)}^{i_1,\pi_{i_1}} \cdot {\bm{S}_{{U}_{1}(1)}^{\pi_{i_1}}}^\top \cdot ( \bm{S}_{{U}_{1}(1)}^{\pi_{i_1}} \cdot {\bm{S}_{{U}_{1}(1)}^{\pi_{i_1}}}^\top)^{-1})^\top$.

\section{Proof of Proposition 6}
The original optimization problem is
\begin{equation*}
\arg \min_{\bm{U}_{1}}~~\alpha||\mathcal{P}_{\Omega}(\mathcal{X} - \mathcal{S} \times_{1} \bm{U}_{1}^\top \times_{2} \bm{U}_{2}^\top \times_{3} \bm{U}_{3}^\top )||_{F}^2 + (1 - \alpha)||\bm{y} - \bm{1}_{M} \cdot \tilde{\beta}_{0} - \bm{S}_{(4)} \cdot \tilde{\bm{\beta}}_{1}||_{F}^{2},
\end{equation*}

\noindent which is equivalent to the following problem when data is incomplete:
\begin{equation*}
\arg \min_{\bm{U}_{1}}~~\alpha||\mathcal{X} - (\mathcal{S} \times_{1} \bm{U}_{1}^\top \times_{2} \bm{U}_{2}^\top \times_{3} \bm{U}_{3}^\top )\odot \text{logic}(\mathcal{X})||_{F}^2 + (1 - \alpha)||\bm{y} - \bm{1}_{M} \cdot \tilde{\beta}_{0} - \bm{S}_{(4)} \cdot \tilde{\bm{\beta}}_{1}||_{F}^{2},
\end{equation*}

\noindent where $\odot$ is the inner product, \text{logic}($\mathcal{X}$) denotes the logical value of $\mathcal{X}$. Since the problem is convex, it can be solved by setting the derivatives to be zeros, i.e., $\frac{d\Psi}{d\bm{U}_{1}} = \mathbf{0}$, where $\Psi = \alpha||\mathcal{X} - (\mathcal{S} \times_{1} \bm{U}_{1}^\top \times_{2} \bm{U}_{2}^\top \times_{3} \bm{U}_{3}^\top )\odot  \text{logic}(\mathcal{X})||_{F}^2 + (1 - \alpha)||\bm{y} - \bm{1}_{M} \cdot \tilde{\beta}_{0} - \bm{S}_{(4)} \cdot \tilde{\bm{\beta}}_{1}||_{F}^{2}$. This implies 
$ \frac{d}{d\bm{U}_{1}}(||\mathcal{X} - (\mathcal{S} \times_{1} \bm{U}_{1}^\top \times_{2} \bm{U}_{2}^\top \times_{3} \bm{U}_{3}^\top) \odot \text{logic}(\mathcal{X}) ||_{F}^2) = \bm{0}$. According to the communication law of tensor mode multiplication, we have $\frac{d}{d\bm{U}_{1}}(||\mathcal{X} - [(\mathcal{S} \times_{2} \bm{U}_{2}^\top \times_{3} \bm{U}_{3}^\top) \times_{1} \bm{U}_{1}^\top] \odot \text{logic}(\mathcal{X}) ||_{F}^2) = \bm{0}$. Thus, $ \frac{d}{d\bm{U}_{1}} (||\mathcal{X} - (\mathcal{S}_{{U}_{1}} \times_{1} \bm{U}_{1}^\top )\odot \text{logic}(\mathcal{X})||_{F}^2)= \bm{0}$, where $\mathcal{S}_{{U}_{1}} = \mathcal{S} \times_{2} \bm{U}_{2}^\top \times_{3} \bm{U}_{3}^\top$. Furthermore, we have $\frac{d}{d\bm{U}_{1}}(||\bm{X}_{(1)} - (\bm{U}_{1}^\top  \cdot \bm{S}_{{U}_{1}(1)})\odot \text{logic}(\bm{X}_{(1)}) ||_{F}^2) = \bm{0}$ due to the fact that $||\mathcal{S}||_{F}^{2} = ||\bm{S}_{(n)}||_{F}^{2}$ and $\mathcal{S} \times_{n} \bm{U} = \bm{U} \cdot \bm{S}_{(n)}$ (a property of tensor mode multiplication).

\begin{figure*}[!htp]
\centering
 \includegraphics[width=0.8\textwidth]{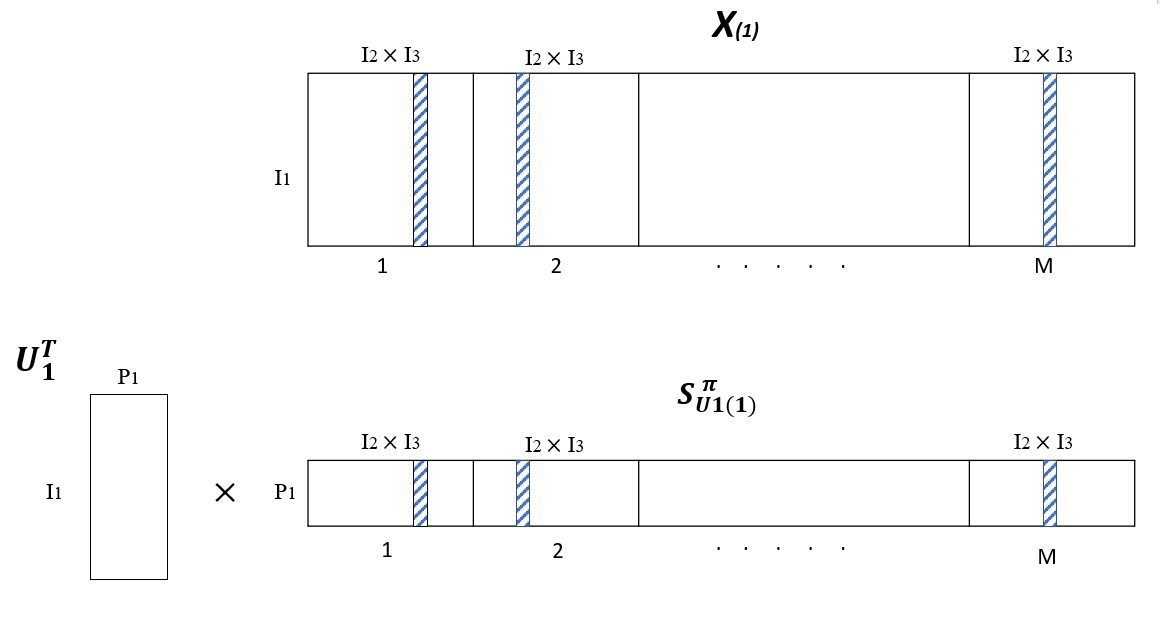}
 \caption{An illustration of the data missing pattern in Proposition 9 (stripes representing available columns).}
 \label{fig: U1 missing}
\end{figure*}

As discussed earlier, for applications with missing images, the indices of tensor $\mathcal{X}$'s missing entries can be denoted as $\Omega \subseteq \lbrace (:, :, i_3, m)$, $1\leq i_3\leq I_3,1\leq m\leq M\}$, where ``$:$" denotes all the indices in a dimension. As a result, it can be easily shown that $\mathcal{X}$'s mode-1 matricization $\bm{X}_{(1)}$ has missing columns (see Figure \ref{fig: U1 missing} for an illustration). Let $\pi$ be the set consisting of the indices of available columns in $\bm{X}_{(1)}$, then we need to solve $ \frac{d}{d\bm{U}_{1}}(||\bm{X}_{(1)}^{\pi} - \bm{U}_{1}^\top  \cdot \bm{S}_{{U}_{1}(1)}^{\pi} ||_{F}^2) = \bm{0}$, where $\bm{S}_{{U}_{1}(1)}^{\pi}$ denotes a matrix constituting the $\pi$ columns of $\bm{S}_{{U}_{1}(1)}$. Thus, we have $2(\bm{X}_{(1)}^{\pi} - \bm{U}_{1}^\top  \cdot \bm{S}_{{U}_{1}(1)}^{\pi}) \cdot (-{\bm{S}_{{U}_{1}(1)}^{\pi}}^\top) = \bm{0}$. Thus, $\bm{U}_{1}^\top \cdot \bm{S}_{{U}_{1}(1)}^{\pi} \cdot {\bm{S}_{{U}_{1}(1)}^{\pi}}^\top = \bm{X}_{(1)}^{\pi} \cdot {\bm{S}_{{U}_{1}(1)}^{\pi}}^\top$, which gives that $ \bm{U}_{1}^\top = \bm{X}_{(1)}^{\pi} \cdot {\bm{S}_{{U}_{1}(1)}^{\pi}}^\top \cdot ( \bm{S}_{{U}_{1}(1)}^{\pi} \cdot {\bm{S}_{{U}_{1}(1)}^{\pi}}^\top)^{-1}$. This yields the solution $\bm{U}_{1} = (\bm{X}_{(1)}^{\pi} \cdot {\bm{S}_{{U}_{1}(1)}^{\pi}}^\top \cdot ( \bm{S}_{{U}_{1}(1)}^{\pi} \cdot {\bm{S}_{{U}_{1}(1)}^{\pi}}^\top)^{-1})^\top$.


\section{Proof of Proposition 7}

The original optimization problem is 
\begin{equation*}
\arg \min_{\bm{U}_{2}}~~\alpha||\mathcal{P}_{\Omega}(\mathcal{X} - \mathcal{S} \times_{1} \bm{U}_{1}^\top \times_{2} \bm{U}_{2}^\top \times_{3} \bm{U}_{3}^\top )||_{F}^2 + (1 - \alpha)||\bm{y} - \bm{1}_{M} \cdot \tilde{\beta}_{0} - \bm{S}_{(4)} \cdot \tilde{\bm{\beta}}_{1}||_{F}^{2},
\end{equation*}

\noindent which is equivalent to the following problem when data is missing:
\begin{equation*}
\arg \min_{\bm{U}_{2}}~~\alpha||\mathcal{X} - (\mathcal{S} \times_{1} \bm{U}_{1}^\top \times_{2} \bm{U}_{2}^\top \times_{3} \bm{U}_{3}^\top )\odot \text{logic}(\mathcal{X})||_{F}^2 + (1 - \alpha)||\bm{y} - \bm{1}_{M} \cdot \tilde{\beta}_{0} - \bm{S}_{(4)} \cdot \tilde{\bm{\beta}}_{1}||_{F}^{2},
\end{equation*}

\noindent where $\odot$ is the inner product, \text{logic}($\mathcal{X}$) denotes the logical value of $\mathcal{X}$. Since the problem is convex, it can be solved by setting the derivatives to be zeros, i.e., $ \frac{d\Psi}{d\bm{U}_{2}} = \bm{0}$,  where $\Psi = \alpha||\mathcal{X} - (\mathcal{S} \times_{1} \bm{U}_{1}^\top \times_{2} \bm{U}_{2}^\top \times_{3} \bm{U}_{3}^\top )\odot \text{logic}(\mathcal{X})||_{F}^2 + (1 - \alpha)||\bm{y} - \bm{1}_{M} \cdot \tilde{\beta}_{0} - \bm{S}_{(4)} \cdot \tilde{\bm{\beta}}_{1}||_{F}^{2}$. This implies 
$ \frac{d}{d\bm{U}_{2}}(||\mathcal{X} - (\mathcal{S} \times_{1} \bm{U}_{1}^\top \times_{2} \bm{U}_{2}^\top \times_{3} \bm{U}_{3}^\top) \odot \text{logic}(\mathcal{X}) ||_{F}^2) = \bm{0}$. According to the communication law of tensor mode multiplication, we have $\frac{d}{d\bm{U}_{2}}(||\mathcal{X} - [(\mathcal{S} \times_{1} \bm{U}_{1}^\top \times_{3} \bm{U}_{3}^\top) \times_{2} \bm{U}_{2}^\top] \odot \text{logic}(\mathcal{X}) ||_{F}^2) = \bm{0}$. Thus, $\frac{d}{d\bm{U}_{2}} (||\mathcal{X} - (\mathcal{S}_{{U}_{2}} \times_{2} \bm{U}_{2}^\top )\odot \text{logic}(\mathcal{X})||_{F}^2)= \bm{0}$, where $\mathcal{S}_{{U}_{2}} = \mathcal{S} \times_{1} \bm{U}_{1}^\top \times_{3} \bm{U}_{3}^\top$. Furthermore, we have $\frac{d}{d\bm{U}_{2}}(||\bm{X}_{(2)} - (\bm{U}_{2}^\top  \cdot \bm{S}_{{U}_{2}(2)})\odot \text{logic}(\bm{X}_{(2)}) ||_{F}^2) = \bm{0}$ since $||\mathcal{S}||_{F}^{2} = ||\bm{S}_{(n)}||_{F}^{2}$ and $\mathcal{S} \times_{n} \bm{U} = \bm{U} \cdot \bm{S}_{(n)}$.

\begin{figure*}[!htp]
\centering
 \includegraphics[width=1\textwidth]{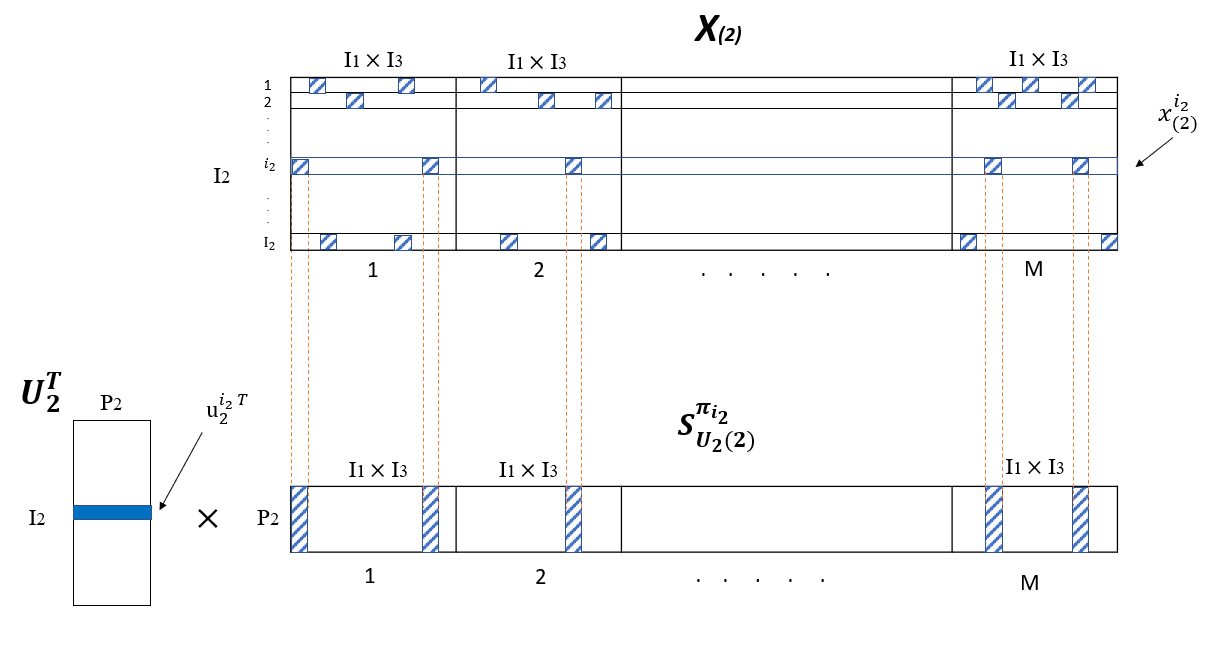}
 \caption{An illustration of the data missing pattern in Proposition 6 (stripes representing available entries).}
 \label{fig: U2 Entry Missing}
\end{figure*}

Figure \ref{fig: U2 Entry Missing} shows the pattern of mode-2 matricization of the 4D tensor $\mathcal{X}$ when it has missing entries whose indices can be denoted by a set $\Omega \subseteq \{ (i_1, i_2, i_3, m),1\leq i_1\leq I_1, 1\leq i_2\leq I_2,1\leq i_3\leq I_3,1\leq m\leq M\}$. Based on Lemma \ref{lemma1}, we can sequentially optimize each column of $\bm{U}_{2}$.

Specifically, we denote the $i_2$th row in $\bm{U}_{2}^\top$ as $\bm{u}_{2}^{i_2 \top}$ (blue solid row of $\bm{U}_{2}^\top$ in Figure \ref{fig: U2 Entry Missing}). The available entries in the $i_2$th row of $\bm{X}_{(2)}$ are denoted as $\bm{x}_{(2)}^{i_2,\pi_{i_2}}$ (blue striped squares of $\bm{x}_{(2)}^{i_2}$ in Figure \ref{fig: U2 Entry Missing}). In $\bm{S}_{U_2(2)}^{\pi_{i_2}}$, we choose the columns whose indices are the same as those of the available entries in $\bm{x}_{(2)}^{i_2}$ (blue striped columns of $\bm{S}_{U_2(2)}^{\pi_{i_2}}$ in Figure \ref{fig: U2 Entry Missing}). As a result, we have $\frac{d}{d\bm{u}_{2}^{i_2}}(\Sigma_{{i_2} = 1}^{I_2}||\bm{x}_{(2)}^{i_2,\pi_{i_2}} - ({\bm{u}_{2}^{i_2 \top}}  \cdot \bm{S}_{{U}_{2}(2)}^{\pi_{i_2}})||_{F}^2) = \bm{0}$. Since on the derivative of $\bm{u}_{2}^{i_2}$ is taken, we have $\frac{d}{d\bm{u}_{2}^{i_2}}(||\bm{x}_{(2)}^{i_2,\pi_{i_2}} - ({\bm{u}_{2}^{i_2 \top}}  \cdot \bm{S}_{{U}_{2}(2)}^{\pi_{i_2}})||_{F}^2) = \bm{0}$. By taking the derivative of the Frobenius norm, we have
$2(\bm{x}_{(2)}^{i_2,\pi_{i_2}} - \bm{u}_{2}^{i_2 \top}  \cdot \bm{S}_{{U}_{2}(2)}^{\pi_{i_2}}) \cdot (-{\bm{S}_{{U}_{2}(2)}^{\pi_{i_2}}}^\top) = \bm{0}$. Thus, $\bm{u}_{2}^{i_2 \top} \cdot \bm{S}_{{U}_{2}(2)}^{\pi_{i_2}} \cdot {\bm{S}_{{U}_{2}(2)}^{\pi_{i_2}}}^\top = \bm{x}_{(2)}^{i_2,\pi_{i_2}} \cdot {\bm{S}_{{U}_{2}(2)}^{\pi_{i_2}}}^\top$, which gives that $\bm{u}_{2}^{i_2} = (\bm{x}_{(2)}^{i_2,\pi_{i_2}} \cdot {\bm{S}_{{U}_{2}(2)}^{\pi_{i_2}}}^\top \cdot ( \bm{S}_{{U}_{2}(2)}^{\pi_{i_2}} \cdot {\bm{S}_{{U}_{2}(2)}^{\pi_{i_2}}}^\top)^{-1})^\top$. 

\section{Proof of Proposition 8}

The original optimization problem is 
\begin{equation*}
\arg \min_{\bm{U}_{2}}~~\alpha||\mathcal{P}_{\Omega}(\mathcal{X} - \mathcal{S} \times_{1} \bm{U}_{1}^\top \times_{2} \bm{U}_{2}^\top \times_{3} \bm{U}_{3}^\top )||_{F}^2 + (1 - \alpha)||\bm{y} - \bm{1}_{M} \cdot \tilde{\beta}_{0} - \bm{S}_{(4)} \cdot \tilde{\bm{\beta}}_{1}||_{F}^{2},
\end{equation*}

\noindent which is equivalent to the following problem when data is incomplete:
\begin{equation*}
\arg \min_{\bm{U}_{2}}~~\alpha||\mathcal{X} - (\mathcal{S} \times_{1} \bm{U}_{1}^\top \times_{2} \bm{U}_{2}^\top \times_{3} \bm{U}_{3}^\top )\odot \text{logic}(\mathcal{X})||_{F}^2 + (1 - \alpha)||\bm{y} - \bm{1}_{M} \cdot \tilde{\beta}_{0} - \bm{S}_{(4)} \cdot \tilde{\bm{\beta}}_{1}||_{F}^{2},
\end{equation*}

\noindent where $\odot$ is the inner product, \text{logic}($\mathcal{X}$) denotes the logical value of $\mathcal{X}$. Since the optimization criterion is convex, it can be solved by setting the derivatives to be zeros, i.e., $ \frac{d\Psi}{d\bm{U}_{2}} = \mathbf{0}$, where $\Psi = \alpha||\mathcal{X} - (\mathcal{S} \times_{1} \bm{U}_{1}^\top \times_{2} \bm{U}_{2}^\top \times_{3} \bm{U}_{3}^\top )\odot \text{logic}(\mathcal{X})||_{F}^2 + (1 - \alpha)||\bm{y} - \bm{1}_{M} \cdot \tilde{\beta}_{0} - \bm{S}_{(4)} \cdot \tilde{\bm{\beta}}_{1}||_{F}^{2}$. This implies 
$\frac{d}{d\bm{U}_{2}}(||\mathcal{X} - (\mathcal{S} \times_{1} \bm{U}_{1}^\top \times_{2} \bm{U}_{2}^\top \times_{3} \bm{U}_{3}^\top) \odot \text{logic}(\mathcal{X}) ||_{F}^2) = \bm{0}$. According to the communication law of tensor mode multiplication, we have $\frac{d}{d\bm{U}_{2}}(||\mathcal{X} - [(\mathcal{S} \times_{1} \bm{U}_{1}^\top \times_{3} \bm{U}_{3}^\top) \times_{2} \bm{U}_{2}^\top] \odot \text{logic}(\mathcal{X}) ||_{F}^2) = \bm{0}$.
As a result, $\frac{d}{d\bm{U}_{2}} (||\mathcal{X} - (\mathcal{S}_{{U}_{2}} \times_{2} \bm{U}_{2}^\top )\odot \text{logic}(\mathcal{X})||_{F}^2)= \bm{0}$, where $\mathcal{S}_{{U}_{2}} = \mathcal{S} \times_{1} \bm{U}_{1}^\top \times_{3} \bm{U}_{3}^\top$. Furthermore, we have $\frac{d}{d\bm{U}_{2}}(||\bm{X}_{(2)} - (\bm{U}_{2}^\top  \cdot \bm{S}_{{U}_{2}(2)})\odot \text{logic}(\bm{X}_{(2)}) ||_{F}^2) = \bm{0}$ since $||\mathcal{S}||_{F}^{2} = ||\bm{S}_{(n)}||_{F}^{2}$ and $\mathcal{S} \times_{n} \bm{U} = \bm{U} \cdot \bm{S}_{(n)}$.

\begin{figure*}[!htp]
\centering
 \includegraphics[width=0.8\textwidth]{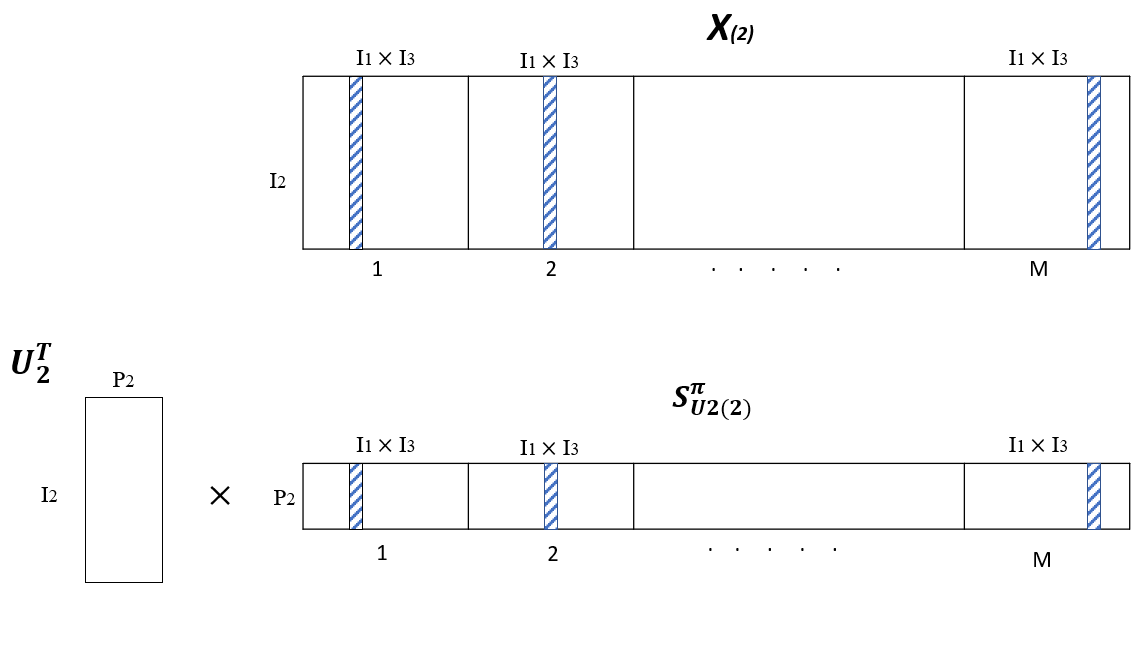}
 \caption{An illustration of the data missing pattern in Proposition 10 (stripes representing available columns).}
 \label{fig: U2 missing}
\end{figure*}

It can be easily shown that $\mathcal{X}$'s mode-2 matricization $\bm{X}_{(2)}$ has missing columns as well (see Figure \ref{fig: U2 missing} for an illustration). Therefore, similar to the proof of Proposition 9, we have $\frac{d}{d\bm{U}_{2}}(||\bm{X}_{(2)}^{\pi} - \bm{U}_{2}^\top  \cdot \bm{S}_{{U}_{2}(2)}^{\pi} ||_{F}^2) = \bm{0}$, where $\pi$ denotes the indices of available columns in $\bm{X}_{(2)}$, $\bm{S}_{{U}_{2}(2)}^{\pi}$ denotes a matrix constituting the $\pi$ columns of $\bm{S}_{{U}_{2}(2)}$. As a result, we have $2(\bm{X}_{(2)}^{\pi} - \bm{U}_{2}^\top  \cdot \bm{S}_{{U}_{2}(2)}^{\pi}) \cdot (-{\bm{S}_{{U}_{2}(2)}^{\pi}}^\top) = \bm{0}$. Thus, $\bm{U}_{2}^\top \cdot \bm{S}_{{U}_{2}(2)}^{\pi} \cdot {\bm{S}_{{U}_{2}(2)}^{\pi}}^\top = \bm{X}_{(2)}^{\pi} \cdot {\bm{S}_{{U}_{2}(2)}^{\pi}}^\top$, which gives that $\bm{U}_{2}^\top = \bm{X}_{(2)}^{\pi} \cdot {\bm{S}_{{U}_{2}(2)}^{\pi}}^\top \cdot ( \bm{S}_{{U}_{2}(2)}^{\pi} \cdot {\bm{S}_{{U}_{2}(2)}^{\pi}}^\top)^{-1}$. This yields the analytical solution $\bm{U}_{2} = (\bm{X}_{(2)}^{\pi} \cdot {\bm{S}_{{U}_{2}(2)}^{\pi}}^\top \cdot ( \bm{S}_{{U}_{2}(2)}^{\pi} \cdot {\bm{S}_{{U}_{2}(2)}^{\pi}}^\top)^{-1})^\top$.

\section{Proof of Proposition 9}

The original optimization problem is 
\begin{equation*}
\arg \min_{\bm{U}_{3}}~~\alpha||\mathcal{P}_{\Omega}(\mathcal{X} - \mathcal{S} \times_{1} \bm{U}_{1}^\top \times_{2} \bm{U}_{2}^\top \times_{3} \bm{U}_{3}^\top )||_{F}^2 + (1 - \alpha)||\bm{y} - \bm{1}_{M} \cdot \tilde{\beta}_{0} - \bm{S}_{(4)} \cdot \tilde{\bm{\beta}}_{1}||_{F}^{2},
\end{equation*}

\noindent which is equivalent to the following problem when data is missing:
\begin{equation*}
\arg \min_{\bm{U}_{3}}~~\alpha||\mathcal{X} - (\mathcal{S} \times_{1} \bm{U}_{1}^\top \times_{2} \bm{U}_{2}^\top \times_{3} \bm{U}_{3}^\top )\odot \text{logic}(\mathcal{X})||_{F}^2 + (1 - \alpha)||\bm{y} - \bm{1}_{M} \cdot \tilde{\beta}_{0} - \bm{S}_{(4)} \cdot \tilde{\bm{\beta}}_{1}||_{F}^{2},
\end{equation*}

\noindent where $\odot$ is the inner product, \text{logic}($\mathcal{X}$) denotes the logical value of $\mathcal{X}$. Since the problem is convex, it can be solved by setting the derivatives to be zeros, i.e., $ \frac{d\Psi}{d\bm{U}_{3}} = \bm{0}$,  where $\Psi = \alpha||\mathcal{X} - (\mathcal{S} \times_{1} \bm{U}_{1}^\top \times_{2} \bm{U}_{2}^\top \times_{3} \bm{U}_{3}^\top )\odot \text{logic}(\mathcal{X})||_{F}^2 + (1 - \alpha)||\bm{y} - \bm{1}_{M} \cdot \tilde{\beta}_{0} - \bm{S}_{(4)} \cdot \tilde{\bm{\beta}}_{1}||_{F}^{2}$. This implies 
$ \frac{d}{d\bm{U}_{3}}(||\mathcal{X} - (\mathcal{S} \times_{1} \bm{U}_{1}^\top \times_{2} \bm{U}_{2}^\top \times_{3} \bm{U}_{3}^\top) \odot \text{logic}(\mathcal{X}) ||_{F}^2) = \bm{0}$. According to the communication law of tensor mode multiplication, we have $\frac{d}{d\bm{U}_{3}}(||\mathcal{X} - [(\mathcal{S} \times_{1} \bm{U}_{1}^\top \times_{2} \bm{U}_{2}^\top) \times_{3} \bm{U}_{3}^\top] \odot \text{logic}(\mathcal{X}) ||_{F}^2) = \bm{0}$. Thus, $\frac{d}{d\bm{U}_{3}} (||\mathcal{X} - (\mathcal{S}_{{U}_{3}} \times_{3} \bm{U}_{3}^\top )\odot \text{logic}(\mathcal{X})||_{F}^2)= \bm{0}$, where $\mathcal{S}_{{U}_{3}} = \mathcal{S} \times_{1} \bm{U}_{1}^\top \times_{2} \bm{U}_{2}^\top$. Furthermore, we have $\frac{d}{d\bm{U}_{3}}(||\bm{X}_{(3)} - (\bm{U}_{3}^\top  \cdot \bm{S}_{{U}_{3}(3)})\odot \text{logic}(\bm{X}_{(3)}) ||_{F}^2) = \bm{0}$ since $||\mathcal{S}||_{F}^{2} = ||\bm{S}_{(n)}||_{F}^{2}$ and $\mathcal{S} \times_{n} \bm{U} = \bm{U} \cdot \bm{S}_{(n)}$.

\begin{figure*}[!htp]
\centering
 \includegraphics[width=1\textwidth]{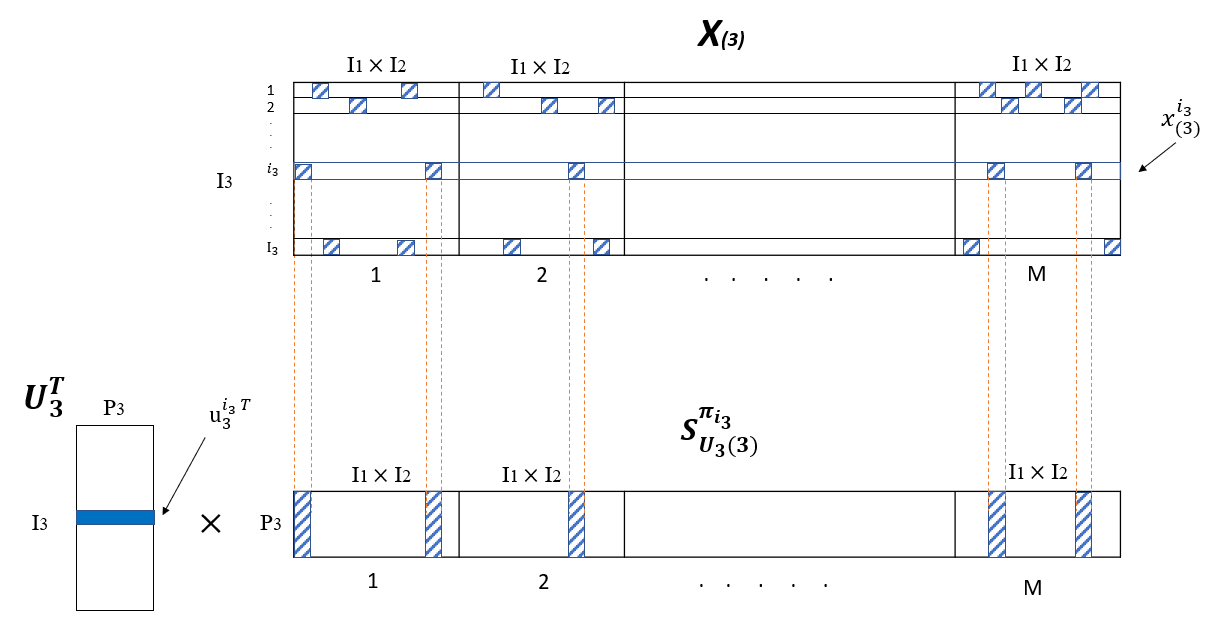}
 \caption{An illustration of the data missing pattern in Proposition 7 (stripes representing available entries).}
 \label{fig: U3 Entry Missing}
\end{figure*}

Figure \ref{fig: U3 Entry Missing} shows the pattern of mode-3 matricization of the 4D tensor $\mathcal{X}$ when it has missing entries whose indices can be denoted by a set $\Omega \subseteq \{ (i_1, i_2, i_3, m),1\leq i_1\leq I_1, 1\leq i_2\leq I_2,1\leq i_3\leq I_3,1\leq m\leq M\}$. Based on Lemma \ref{lemma1}, we can sequentially optimize each column of $\bm{U}_{3}$. The $i_3$th row in $\bm{U}_{3}^\top$ is denoted as $\bm{u}_{3}^{i_3 \top}$ (blue solid row of $\bm{U}_{3}^\top$ in Figure \ref{fig: U3 Entry Missing}). The available entries in the $i_3$th row of $\bm{X}_{(3)}$ are denoted as $\bm{x}_{(3)}^{i_3,\pi_{i_3}}$ (blue striped squares of $\bm{x}_{(3)}^{i_3}$ in Figure \ref{fig: U3 Entry Missing}). In $\bm{S}_{U_3(3)}^{\pi_{i_3}}$, we choose the columns whose indices are the same as those of the available entries of $\bm{x}_{(3)}^{i_3}$ (blue striped columns of $\bm{S}_{U_3(3)}^{\pi_{i_3}}$ in Figure \ref{fig: U3 Entry Missing}). Thus, we have $\frac{d}{d\bm{u}_{3}^{i_3}}(\Sigma_{{i_3} = 1}^{I_3}||\bm{x}_{(3)}^{i_3,\pi_{i_3}} - ({\bm{u}_{3}^{i_3 \top}}  \cdot \bm{S}_{{U}_{3}(3)}^{\pi_{i_3}})||_{F}^2) = \bm{0}$, which yields $\frac{d}{d\bm{u}_{3}^{i_3}}(||\bm{x}_{(3)}^{i_3,\pi_{i_3}} - ({\bm{u}_{3}^{i_3 \top}}  \cdot \bm{S}_{{U}_{3}(3)}^{\pi_{i_3}})||_{F}^2) = \bm{0}$. By taking the derivative of the Frobenius norm, we have
$2(\bm{x}_{(3)}^{i_3,\pi_{i_3}} - \bm{u}_{3}^{i_3 \top}  \cdot \bm{S}_{{U}_{3}(3)}^{\pi_{i_3}}) \cdot (-{\bm{S}_{{U}_{3}(3)}^{\pi_{i_3}}}^\top) = \bm{0}$. Thus, $\bm{u}_{3}^{i_3 \top} \cdot \bm{S}_{{U}_{3}(3)}^{\pi_{i_3}} \cdot {\bm{S}_{{U}_{3}(3)}^{\pi_{i_3}}}^\top = \bm{x}_{(3)}^{i_3,\pi_{i_3}} \cdot {\bm{S}_{{U}_{3}(3)}^{\pi_{i_3}}}^\top$, which gives that $\bm{u}_{3}^{i_3} = (\bm{x}_{(3)}^{i_3,\pi_{i_3}} \cdot {\bm{S}_{{U}_{3}(3)}^{\pi_{i_3}}}^\top \cdot ( \bm{S}_{{U}_{3}(3)}^{\pi_{i_3}} \cdot {\bm{S}_{{U}_{3}(3)}^{\pi_{i_3}}}^\top)^{-1})^\top$.

\section{Proof of Proposition 10}
The original optimization problem is
\begin{equation*}
\arg \min_{S}~~\alpha||\mathcal{P}_{\Omega}(\mathcal{X} - \mathcal{S} \times_{1} \bm{U}_{1}^\top \times_{2} \bm{U}_{2}^\top \times_{3} \bm{U}_{3}^\top )||_{F}^2 + (1 - \alpha)||\bm{y} - \bm{1}_{M} \cdot \beta_{0} - \bm{S}_{(4)} \cdot \bm{\beta}_{1}||_{F}^{2}
\end{equation*}

\noindent which is equivalent to the following problem when data is missing:
\begin{equation*}
\arg \min_{S}~~\alpha||\mathcal{X} - (\mathcal{S} \times_{1} \bm{U}_{1}^\top \times_{2} \bm{U}_{2}^\top \times_{3} \bm{U}_{3}^\top) \odot \text{logic}(\mathcal{X}) ||_{F}^2 + (1 - \alpha)||\bm{y} - \bm{1}_{M} \cdot \beta_{0} - \bm{S}_{(4)} \cdot \bm{\beta}_{1}||_{F}^{2}
\end{equation*}

\noindent where $\odot$ is the inner product, \text{logic}($\mathcal{X}$) denotes the logical value of $\mathcal{X}$. Since the problem is convex, it can be solved by setting the derivative to be zeros--that is $\frac{d\Psi}{d\bm{S}} = \bm{0}$, where $\Psi = \alpha||\mathcal{X} - (\mathcal{S} \times_{1} \bm{U}_{1}^\top \times_{2} \bm{U}_{2}^\top \times_{3} \bm{U}_{3}^\top) \odot \text{logic}(\mathcal{X}) ||_{F}^2 + (1 - \alpha)||\bm{y} - \bm{1}_{M} \cdot \beta_{0} - \bm{S}_{(4)} \cdot \bm{\beta}_{1}||_{F}^{2}$. According to the connection between Kronecker product and tensor mode multiplication, we have 
$\frac{d}{d\bm{S}}(\alpha||\bm{X}_{(4)} - [\bm{S}_{(4)} \cdot (\bm{U}_{3} \otimes \bm{U}_{2} \otimes \bm{U}_{1})] \odot \text{logic}(\bm{X}_{(4)}) ||_{F}^2 + (1 - \alpha)||\bm{y} - \bm{1}_{M} \odot\beta_{0} - \bm{S}_{(4)}  \odot \bm{\beta}_{1}||_{F}^{2}) = \bm{0}$.

\begin{figure*}[!htp]
\centering
 \includegraphics[width=1\textwidth]{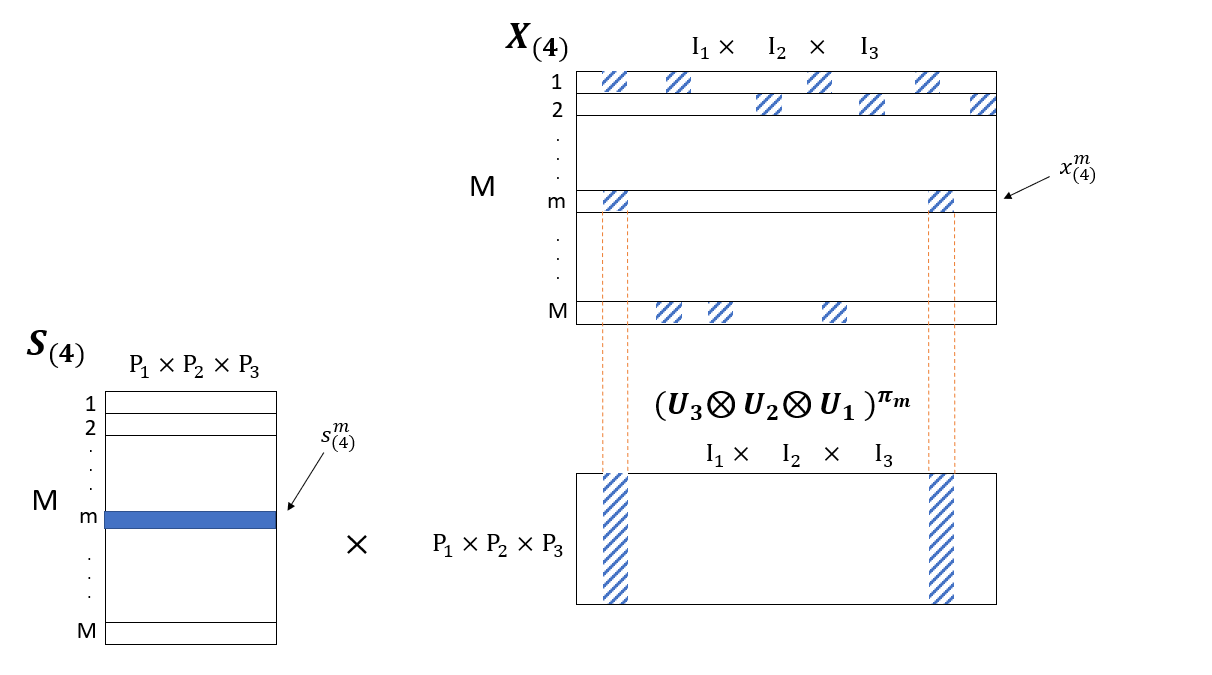}
 \caption{An illustration of the data missing pattern in Proposition 8 (stripes representing available entries).}
 \label{fig: S Entry Missing}
\end{figure*}

Figure \ref{fig: S Entry Missing} shows the pattern of mode-4 matricization of the 4D tensor $\mathcal{X}$ when it has missing entries whose indices can be denoted by a set $\Omega \subseteq \{ (i_1, i_2, i_3, m),1\leq i_1\leq I_1, 1\leq i_2\leq I_2,1\leq i_3\leq I_3,1\leq m\leq M\}$. Based on Lemma \ref{lemma1}, we can sequentially optimize each column of $\bm{U}_{3}$.

The $m$th row in $\bm{S}_{(4)}$ is denoted as $\bm{s}_{(4)}^{m}$ (blue solid row of $\bm{S}_{(4)}$ in Figure \ref{fig: S Entry Missing}). The available entries in the $m$th row of $\bm{X}_{(4)}$ are denoted as $\bm{x}_{(4)}^{m,\pi_{m}}$ (blue striped squares of $\bm{x}_{(4)}^{m}$ in Figure \ref{fig: S Entry Missing}). In $(\bm{U}_{3} \otimes \bm{U}_{2} \otimes \bm{U}_{1})$, we choose the columns whose indices are the same as those of the available entries of $\bm{x}_{(4)}^{m}$ (blue striped columns of $(\bm{U}_{3} \otimes \bm{U}_{2} \otimes \bm{U}_{1})^{\pi_m}$ in Figure \ref{fig: S Entry Missing}). Thus, we have $\frac{d}{d\bm{s}_{(4)}^{m}}(\alpha\lbrace\Sigma_{m = 1}^{M}||\bm{x}_{(4)}^{m,\pi_{m}} - [\bm{s}_{(4)}^{m} \cdot (\bm{U}_{3} \otimes \bm{U}_{2} \otimes \bm{U}_{1})^{\pi_m}] \rbrace ||_{F}^2 + (1 - \alpha)\Sigma_{m = 1}^{M}||y_{m} - \beta_{0} - \bm{s}_{(4)}^{m} \cdot \bm{\beta}_{1}||_{F}^{2}) = \bm{0}$, which yields $\frac{d}{d\bm{s}_{(4)}^{m}}(\alpha\lbrace||\bm{x}_{(4)}^{m,\pi_{m}} - [\bm{s}_{(4)}^{m} \cdot (\bm{U}_{3} \otimes \bm{U}_{2} \otimes \bm{U}_{1})^{\pi_m}] \rbrace ||_{F}^2 + (1 - \alpha)||y_m - \beta_{0} - \bm{s}_{(4)}^{m} \cdot \bm{\beta}_{1}||_{F}^{2}) = \bm{0}$. By taking the derivative of Frobenius norm, we have $2\alpha \cdot [\bm{x}_{(4)}^{m,\pi_{m}} - \bm{s}_{(4)}^{m} \cdot (\bm{U}_{3} \otimes \bm{U}_{2} \otimes \bm{U}_{1})^{\pi_m}]\cdot [- (\bm{U}_{3} \otimes \bm{U}_{2} \otimes \bm{U}_{1})^{\pi_m \top}] + 2(1 - \alpha) \cdot [y_m - \beta_{0} - \bm{s}_{(4)}^{m} \cdot \bm{\beta}_{1}] \cdot (-\bm{\beta}_{1}^\top) = \bm{0}$. Thus, 
$-2\alpha \cdot \bm{x}_{(4)}^{m,\pi_{m}} \cdot (\bm{U}_{3} \otimes \bm{U}_{2} \otimes \bm{U}_{1})^{\pi_m \top} + 2\alpha \cdot \bm{s}_{(4)}^{m} \cdot (\bm{U}_{3} \otimes \bm{U}_{2} \otimes \bm{U}_{1})^{\pi_m} \cdot (\bm{U}_{3} \otimes \bm{U}_{2} \otimes \bm{U}_{1})^{\pi_m \top} + 2(1-\alpha) (y_m - \beta_{0})\cdot (-\bm{\beta}_{1}^\top) + 2(1 - \alpha) \cdot (\bm{s}_{(4)}^{m} \cdot \bm{\beta}_{1} \cdot \bm{\beta}_{1}^\top) = \bm{0}$, which gives that $\bm{s}_{(4)}^{m} = \big[\alpha \cdot \bm{x}_{(4)}^{m,\pi_{m}} \cdot (\bm{U}_{3} \otimes \bm{U}_{2} \otimes \bm{U}_{1})^{\pi_m \top} + (1 - \alpha) \cdot (y_{m} - \beta_{0}) \cdot \bm{\beta}_{1}^\top \big] \cdot \big[ \alpha \cdot (\bm{U}_{3} \otimes \bm{U}_{2} \otimes \bm{U}_{1})^{\pi_m} \cdot (\bm{U}_{3} \otimes \bm{U}_{2} \otimes \bm{U}_{1})^{\pi_m \top} + (1 - \alpha) \cdot \bm{\beta}_{1} \cdot \bm{\beta}_{1}^\top \big]^{-1}$.

\bibliographystyle{chicago}
\bibliography{Bibliography-MM-MC.bib}



\end{document}